\documentclass[sigconf,nonacm]{acmart}
\usepackage{arydshln}
\usepackage{diagbox}
\usepackage{algorithmic}
\usepackage{algorithm}
\usepackage{multirow}
\usepackage{multicol}
\usepackage{adjustbox}
\usepackage{booktabs}
\usepackage{cleveref}
\usepackage{subfig}
\usepackage{balance}

\usepackage{enumitem}
\usepackage{xcolor}

\usepackage{xspace}
\makeatletter
\DeclareRobustCommand\onedot{\futurelet\@let@token\@onedot}
\def\@onedot{\ifx\@let@token.\else.\null\fi\xspace}
\DeclareRobustCommand\nodot{\futurelet\@let@token\@nodot}
\def\@nodot{\ifx\@let@token.\else~\null\fi\xspace}

\makeatother

\AtBeginDocument{%
  }

\begin{document}
\newcommand{\name}{\textsf{P-GSVC}}
\title{{\name}: Layered Progressive 2D Gaussian Splatting \\for Scalable Image and Video}
\renewcommand{\shorttitle}{{\name}: Layered Progressive 2D Gaussian Splatting for Scalable Image and Video}

\author{Longan Wang}
\affiliation{%
  \institution{National University of Singapore}
  \city{Singapore}\country{}}
\email{wanglongan@u.nus.edu}
\orcid{0009-0001-9989-3739}

\author{Yuang Shi}
\affiliation{%
  \institution{National University of Singapore}
  \city{Singapore}\country{}}
\email{yuangshi@u.nus.edu}
\orcid{0000-0002-7893-8512}

\author{Wei Tsang Ooi}
\affiliation{%
  \institution{National University of Singapore}
  \city{Singapore}\country{}}
\email{dcsooiwt@nus.edu.sg}
\orcid{0000-0001-8994-1736}


\begin{abstract}
Gaussian splatting has emerged as a competitive explicit representation for image and video reconstruction.
In this work, we present {\name}, the first layered progressive 2D Gaussian splatting framework that provides a unified solution for scalable Gaussian representation in both images and videos.
{\name} organizes 2D Gaussian splats into a base layer and successive enhancement layers, enabling coarse-to-fine reconstructions.
To effectively optimize this layered representation, we propose a joint training strategy that simultaneously updates Gaussians across layers, aligning their optimization trajectories to ensure inter-layer compatibility and a stable progressive reconstruction.
{\name} supports scalability in terms of both quality and resolution.  Our experiments show that the joint training strategy can gain up to 1.9 dB improvement in PSNR for video and 2.6 dB improvement in PSNR for image when compared to methods that perform sequential layer-wise training. 
Project page: \url{https://longanwang-cs.github.io/PGSVC-webpage/}.
\end{abstract}

\ccsdesc[500]{Information systems~Multimedia streaming}

\keywords{Scalable Image, Scalable Video, Progressive Coding, 2D Gaussian Splatting}


\maketitle

\section{Introduction \label{Sec:Introduction}}

\begin{figure}
\centering
    \begin{minipage}[b]{0.49\textwidth}\centering
      \subfloat[Pruned Gaussian splats]{
        \includegraphics[width=0.48\textwidth]{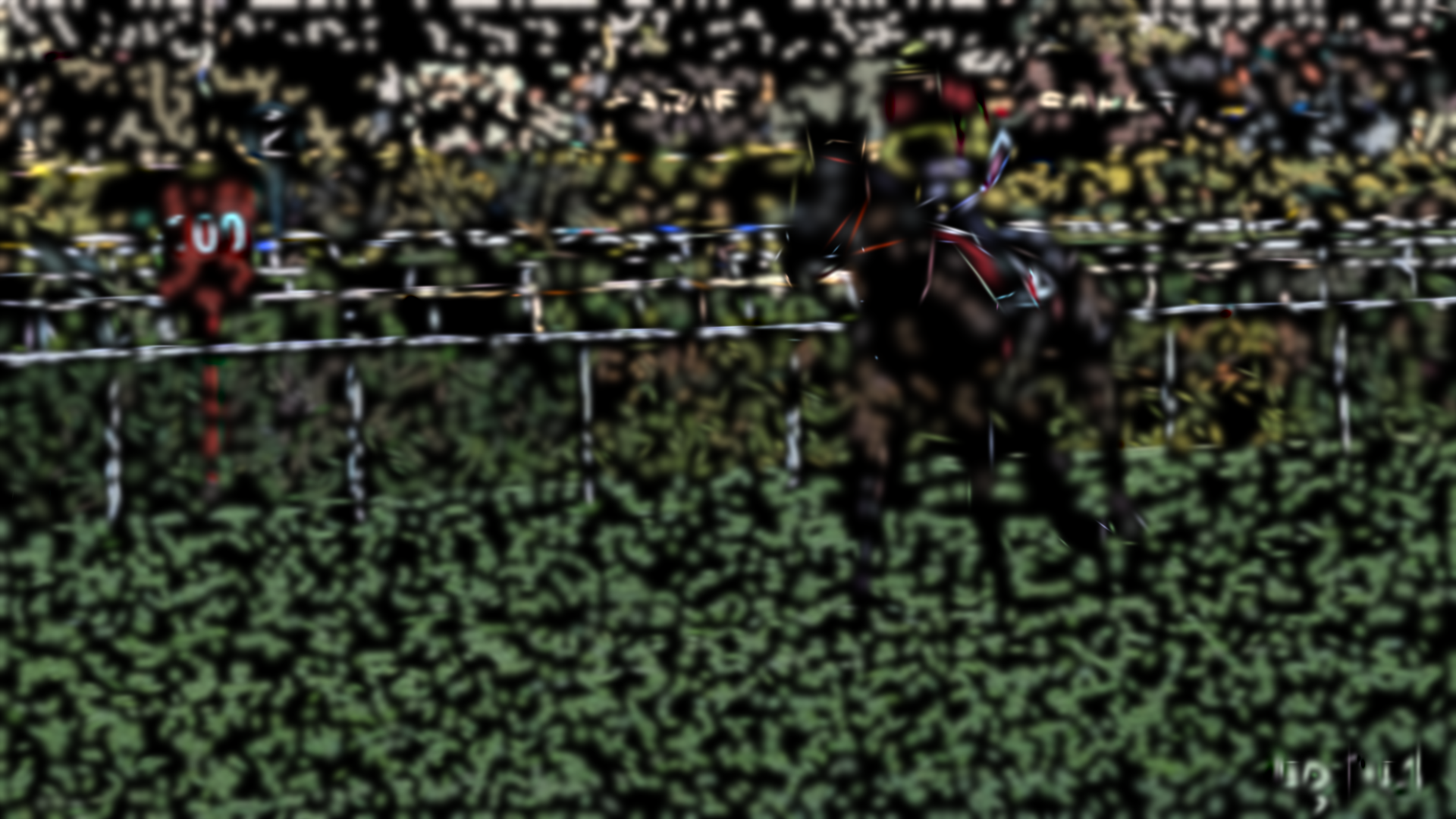}
        \includegraphics[width=0.48\textwidth]{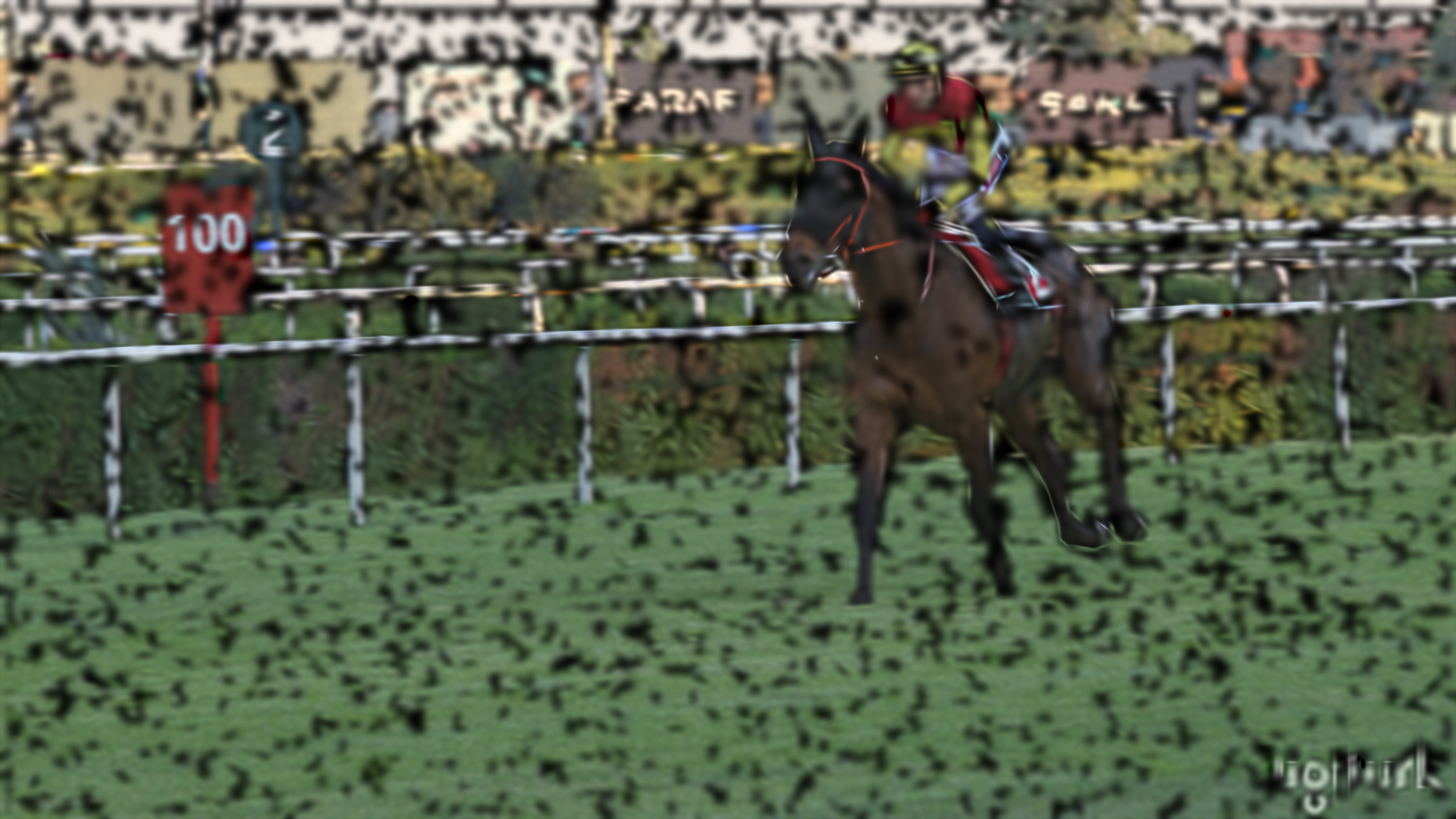}
        \label{fig:unLayered Gaussians}
      }\hfill
    \vspace{0.5em}
    \end{minipage}
    \begin{minipage}[b]{0.49\textwidth}\centering
      \subfloat[P-GSVC]{
        \includegraphics[width=0.48\textwidth]{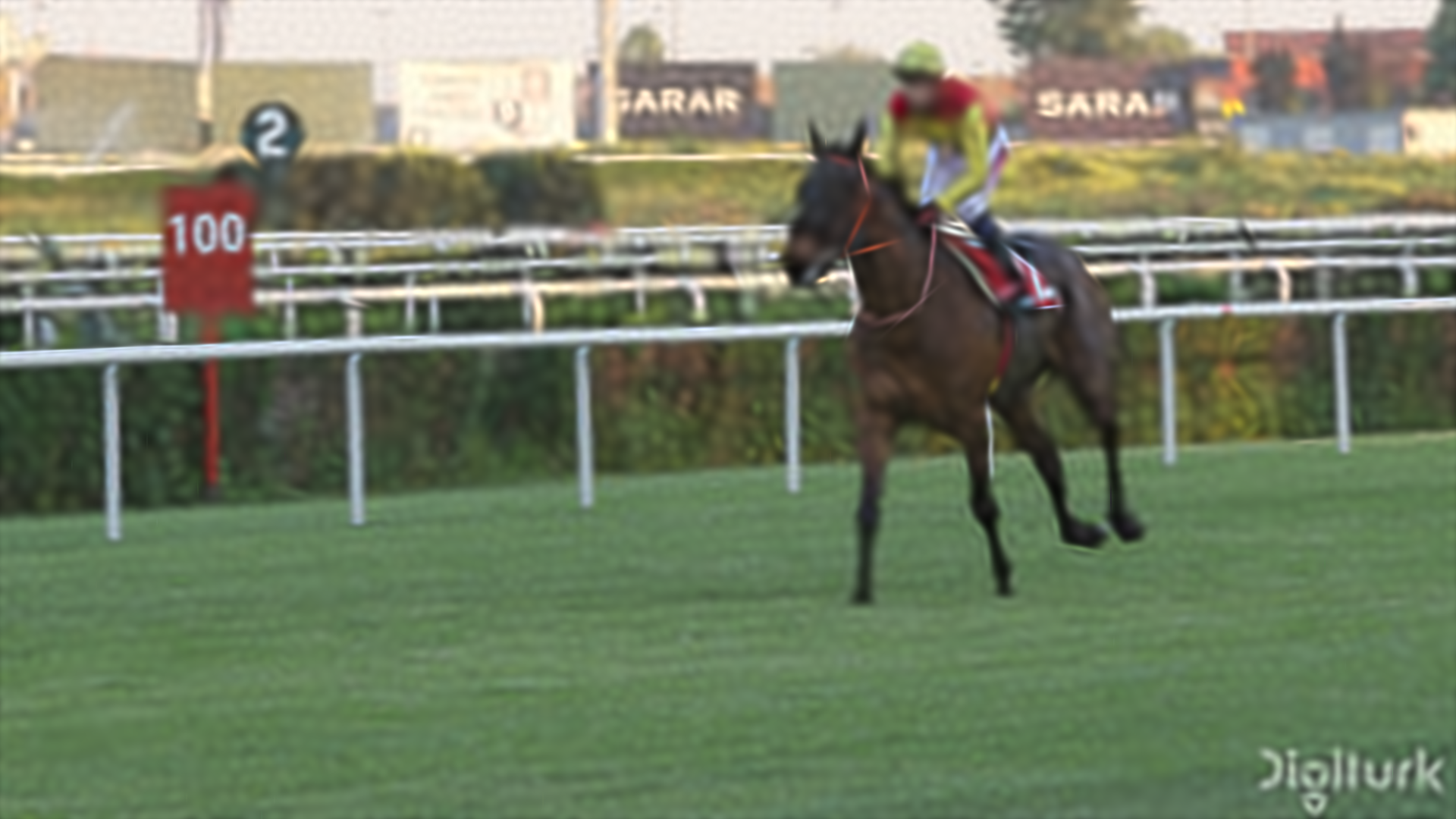}
        \includegraphics[width=0.48\textwidth]{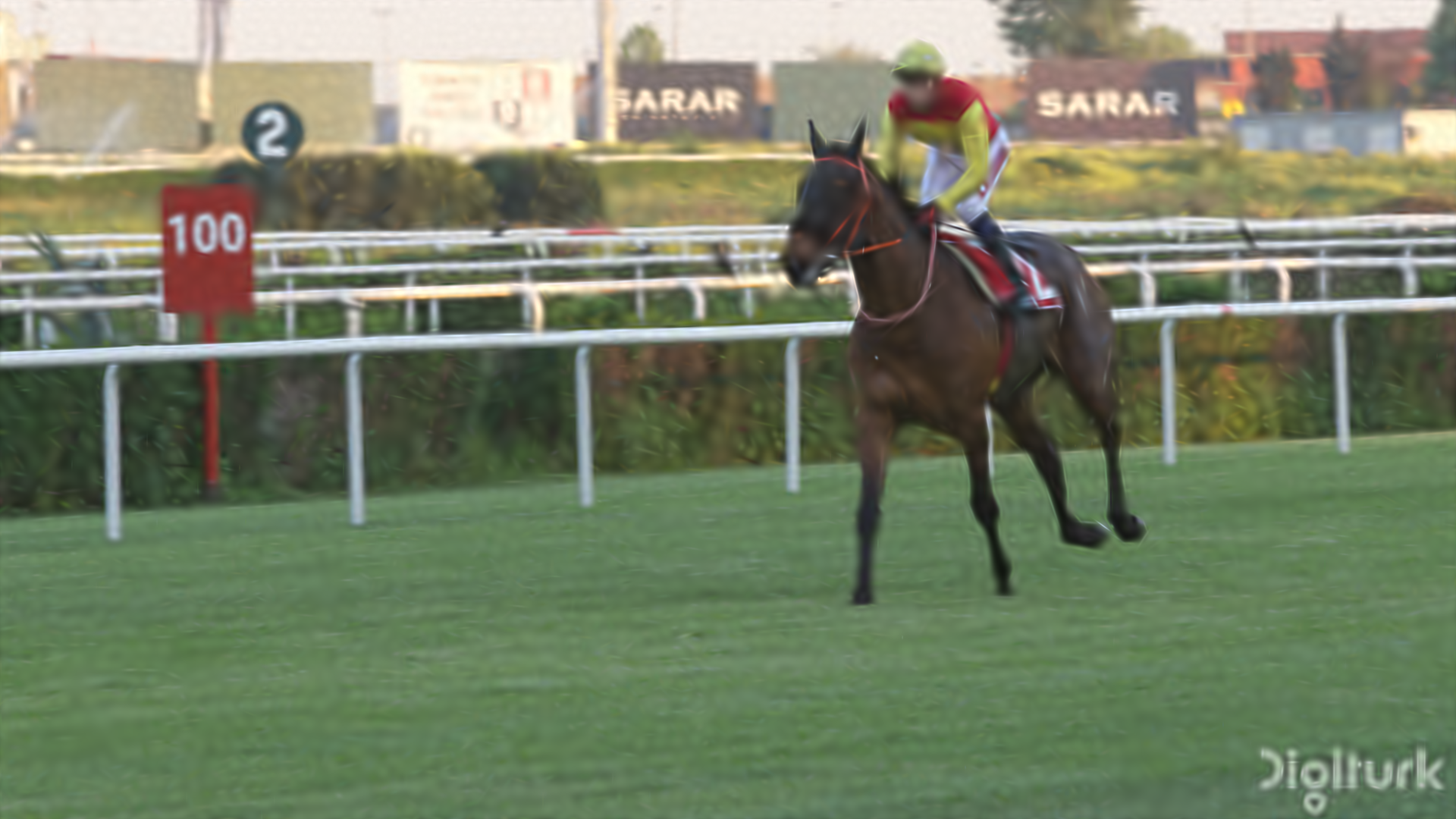}
        \label{fig:Layered Gaussians}
      }\hfill
    \end{minipage}
    \vspace{-1em}
    \caption{Progressive reconstruction of (a) Pruned Gaussian splats and (b) Layered Gaussian splats (our \name). Both columns use the same number of splats, and the quality improves from left to right as more splats are added. Noticeable holes appear in (a), while (b) preserves scene integrity.}
\label{fig:insight-comparision-Layering}
\end{figure}

Scalable image and video codecs support progressive encoding and decoding at multiple bitrates, allowing streaming of videos over fluctuating network bandwidth and displaying on heterogeneous devices.
Progressive codecs have been developed, since three decades ago, to enable such scalability
~\cite{DBLP:books/daglib/0007442, ISO13818-2}.
By organizing the encoded bit streams into a base layer and one or more enhancement layers, scalable coding provides coarse reconstructions from the base layer and progressively higher quality as more layers are decoded.

Classical scalable standards, such as JPEG2000~\cite{DBLP:books/daglib/0007442} for images and HEVC-SHVC~\cite{DBLP:journals/tcsv/BoyceYCR16} for videos, support scalability in resolution and quality.
\textcolor{black}{Recently, learning-based scalable codecs have emerged, replacing hand-crafted modules with end-to-end optimized neural networks and simplifying codec design. 
These methods typically adopt two scalability paradigms:  progressive decoding of hierarchical latent representations~\cite{DBLP:conf/wacv/PrestaTFGC25, DBLP:conf/aaai/LuD0M24, DBLP:conf/mm/WuQHLLYLY24}, or partitioning the neural network parameters into base and enhancement subnetworks to enable quality scalability~\cite{DBLP:journals/tomccap/ZhangZT24, DBLP:conf/mmm/CaoZS24}.}
Such approaches, however, typically require substantial computational resources and data for training.
Furthermore, implicit representations complicate content editing and post-processing, and offer limited fine-grained control over bitrate and quality, which is particularly problematic for adaptive delivery over heterogeneous networks and devices~\cite{DBLP:conf/mmsys/Shi25,DBLP:conf/mmve/ShiZGMO24}.

3D Gaussian Splatting (3DGS)~\cite{DBLP:journals/tog/KerblKLD23} has emerged as a powerful learning-based paradigm for 3D scene representation.
In contrast to neural-based implicit methods such as NeRF~\cite{DBLP:journals/cacm/MildenhallSTBRN22}, 3DGS leverages explicit Gaussian splats as primitives, enabling high-fidelity reconstruction and real-time rendering through differentiable rasterization.
Recently, progressive extensions of 3DGS have been explored, enabling coarse-to-fine reconstruction~\cite{DBLP:conf/cvpr/ZhangZXLX24}, progressive rendering~\cite{DBLP:conf/wacv/Zoomers0MVPM25}, adaptive streaming~\cite{DBLP:conf/3dim/ShiMGO25}, and scalable compression~\cite{DBLP:journals/corr/abs-2503-08511,DBLP:journals/corr/abs-2601-05394}.

The success of Gaussian Splatting in 3D scene reconstruction has motivated the adoption of Gaussian splats as new primitives for representing 2D content, including both images and videos.
For images, 2D Gaussian Splatting (2DGS) has been adopted for efficient image representation~\cite{DBLP:conf/aaai/ZhuLCZJWY25} and compression~\cite{DBLP:conf/eccv/ZhangGXHWQLGZ24}.
For videos, additional compression opportunity arises from temporal redundancy, and new challenges such as dynamic content and scene transitions make video reconstruction more complex.
Prior attempts use 3DGS for editable video reconstruction~\cite{DBLP:conf/nips/Sun0MLC024, DBLP:journals/corr/abs-2411-11024, DBLP:journals/corr/abs-2501-04782}, without a focus on compression.
2DGS has also been proposed for video representation and compression~\cite{DBLP:conf/nossdav/WangSO25, DBLP:conf/cvpr/LeeCL25} as a more lightweight alternative to 3DGS.
Notably, GSVC~\cite{DBLP:conf/nossdav/WangSO25} achieves the state-of-the-art quality-size trade-offs with several techniques: (i) consecutive learning, where Gaussian splats of each P-frame are initialized from the previous frame to exploit temporal redundancy; (ii) Gaussian Splat Pruning (GSP), which removes splats with little contribution to visual quality, reducing bitrate while maintaining fidelity; (iii) Gaussian Splat Augmentation (GSA), which injects new splats to capture dynamic content such as fast motion or new objects; and (iv) Dynamic Key-frame Selection (DKS), which detects scene changes and inserts new I-frames when needed. Together, these mechanisms allow GSVC to balance compression efficiency and adaptability to dynamic scenes, yielding competitive rate–distortion performance when compared to modern open-sourced implementations of codecs such as AV1 and AVC.

While 2DGS is an effective representation for both image and video, extending it to scalable coding is non-trivial.
Since an image can be regarded as a single-frame video, we describe the following challenges in terms of videos for generality.

\begin{figure}[t]
\centering
\begin{minipage}[b]{0.49\textwidth}\centering
  \subfloat[Single-layer]{
    \includegraphics[width=0.49\textwidth]{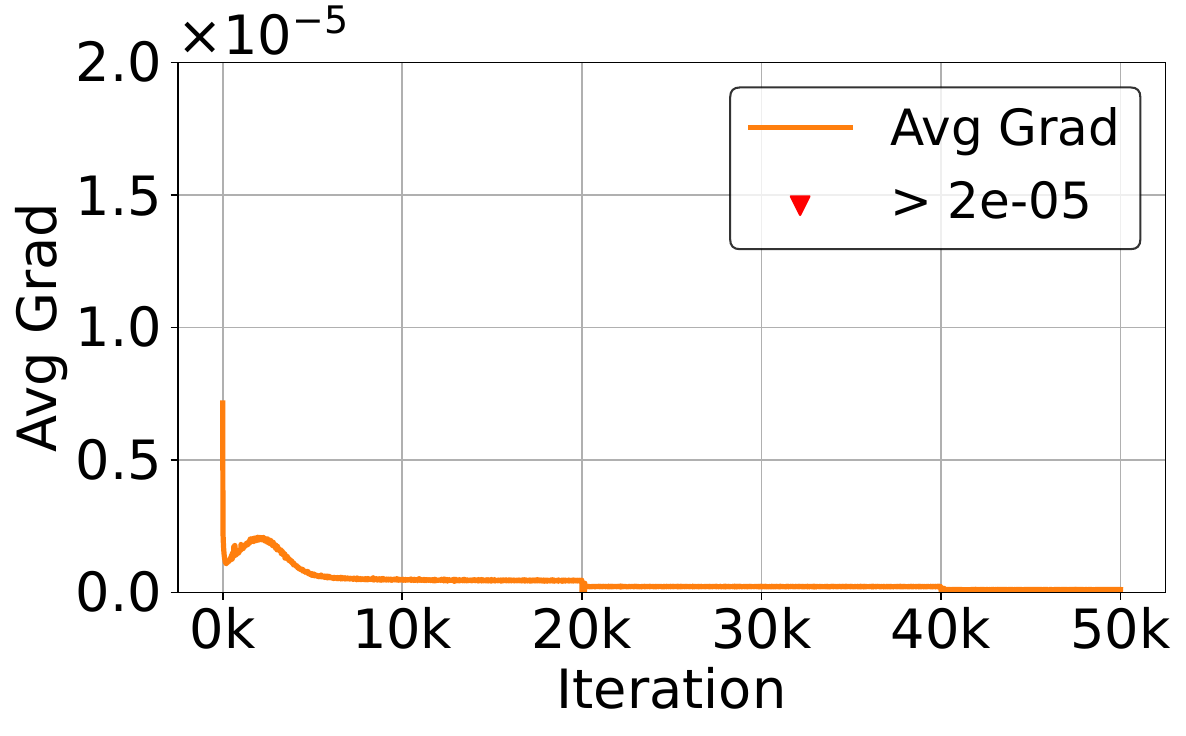}
    \includegraphics[width=0.49\textwidth]{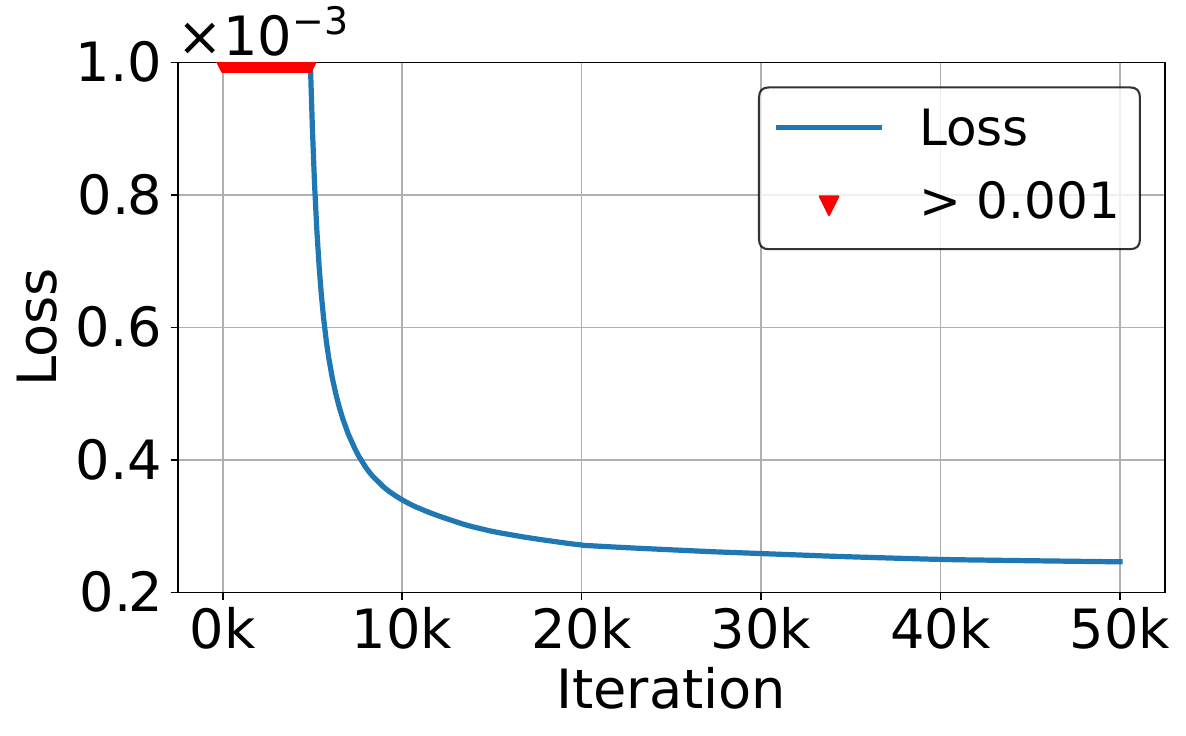}
    \label{fig:single-layer}
  }\hfill
\end{minipage}
\begin{minipage}[b]{0.49\textwidth}\centering
  \subfloat[Multi-layer]{
    \includegraphics[width=0.49\textwidth]{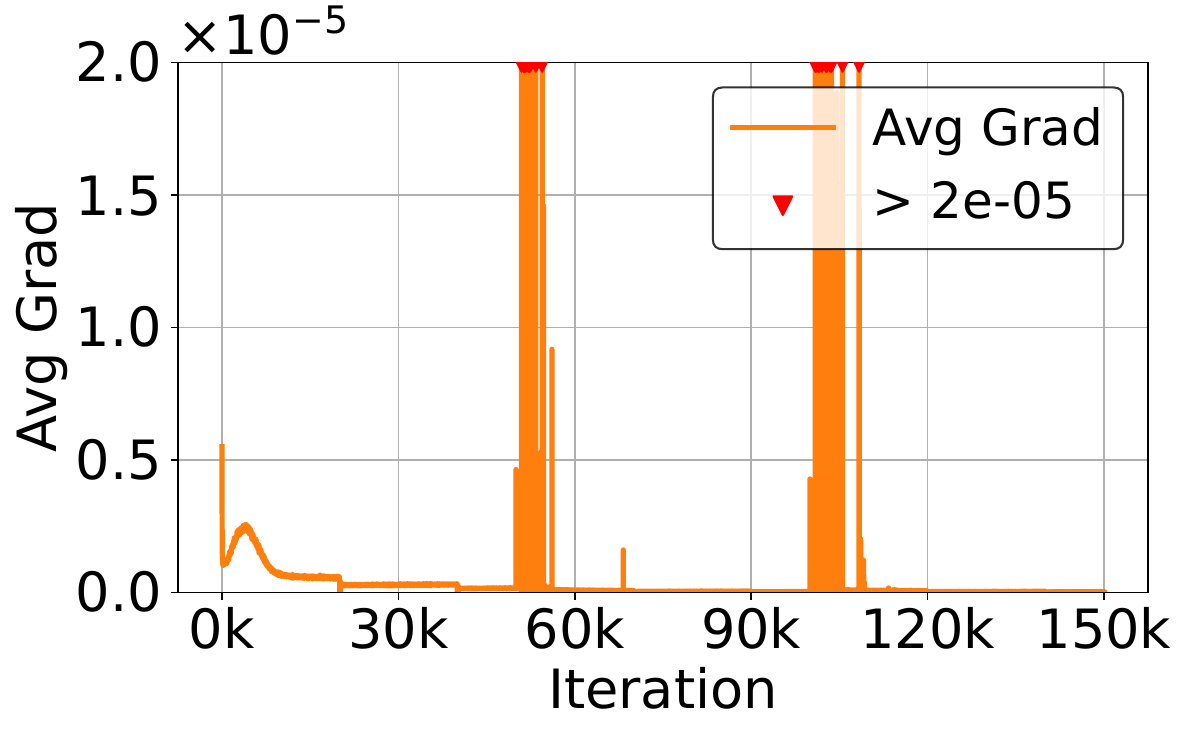}
    \includegraphics[width=0.49\textwidth]{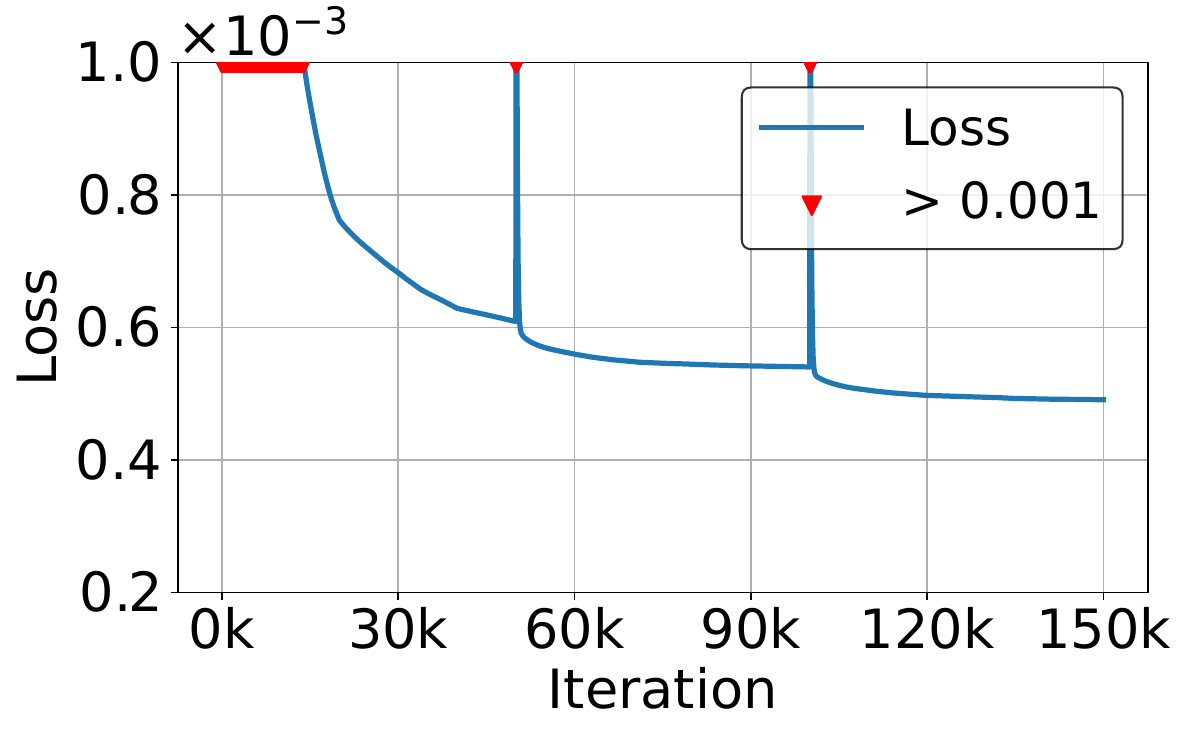}
    \label{fig:sequentially_layer-wise}
  }\hfill
\end{minipage}
\caption{Comparison of gradient and loss over iterations on single-layer and multi-layer optimization strategies: (a) Single-layer optimization in GSVC; (b) Multi-layer trained sequentially in GSVC (the targeted layer is switched at the 50k iteration and the 100k iteration, while the previously trained layers remain frozen). One can observe multi-layer training has issues of unstable convergence and suboptimal local minima.}
\label{fig:insight-comparision}
\end{figure}

To motivate why progressive 2DGS representation is not straightforward, let's consider a naive scheme where we rank the splats by their contribution to the rendered output~\cite{DBLP:conf/nips/FanWWZXW24, DBLP:conf/nossdav/WangSO25} and form the layers based on the contribution: splats with higher contribution form the coarse lower fidelity layers; splats with smaller contribution serve to refine the details and form the higher layers.  
This approach, however, produces visible artifacts at lower layers, such as noticeable gaps in the reconstructed frames, as shown in~\Cref{fig:unLayered Gaussians}. 
The root cause is that splats are jointly trained to overfit the highest fidelity input by design, making them highly interdependent on each other~\cite{DBLP:conf/mmsys/Shi25}. Consequently, removing any subset, even those with small contributions, 
can drastically degrade the overall quality.

Next, consider the approach adopted in 3DGS (e.g., LapisGS~\cite{DBLP:conf/3dim/ShiMGO25}) for resolution scalability, where the layers are trained one-by-one incrementally.  When a new layer is trained, the lower layers that it depends on is fixed (except for opacity).  While this separation of layers can be enforced effectively in static 3D scenes, 
directly extending this strategy to GSVC leads to suboptimal training. Specifically, unlike static representations, videos contain substantial temporal dynamics across frames, including motion, lighting changes, and occlusions. As GSVC uses Gaussian prediction for P-frames, these dynamics increase the interdependency of the Gaussian splats, not only across the layers but also across time.  As a result, constraining different layers to independent objectives leads to conflicting gradients during optimization.
\Cref{fig:sequentially_layer-wise} illustrates such \textit{cross-layer optimization conflict}, showing that a sample multi-layer training under GSVC suffers from unstable convergence and is trapped in suboptimal local minima, with a higher final loss compared to the single-layer baseline in~\Cref{fig:single-layer}. This highlights that optimizing GSVC layer-wise is more difficult than that in static 3D settings.

\Cref{fig:sequentially_layer-wise} also highlights another challenge in layer-wise training.  When we start to train a new enhancement layer at iterations 50k and 100k, the optimization objective is changed and the training exhibits sharp gradient increase.  Eventually, the loss stabilized but with only marginal loss reduction at the enhancement layers. This \textit{progressive instability} indicates that splats learned in lower layers are poorly conditioned as the starting point for training the higher layer, and that the optimization process is sensitive to layer switching. As a result, it struggles to provide high-quality intermediate reconstructions, undermining one of the key requirements for progressive scalability.

\begin{figure}
  \includegraphics[width=0.99\linewidth]{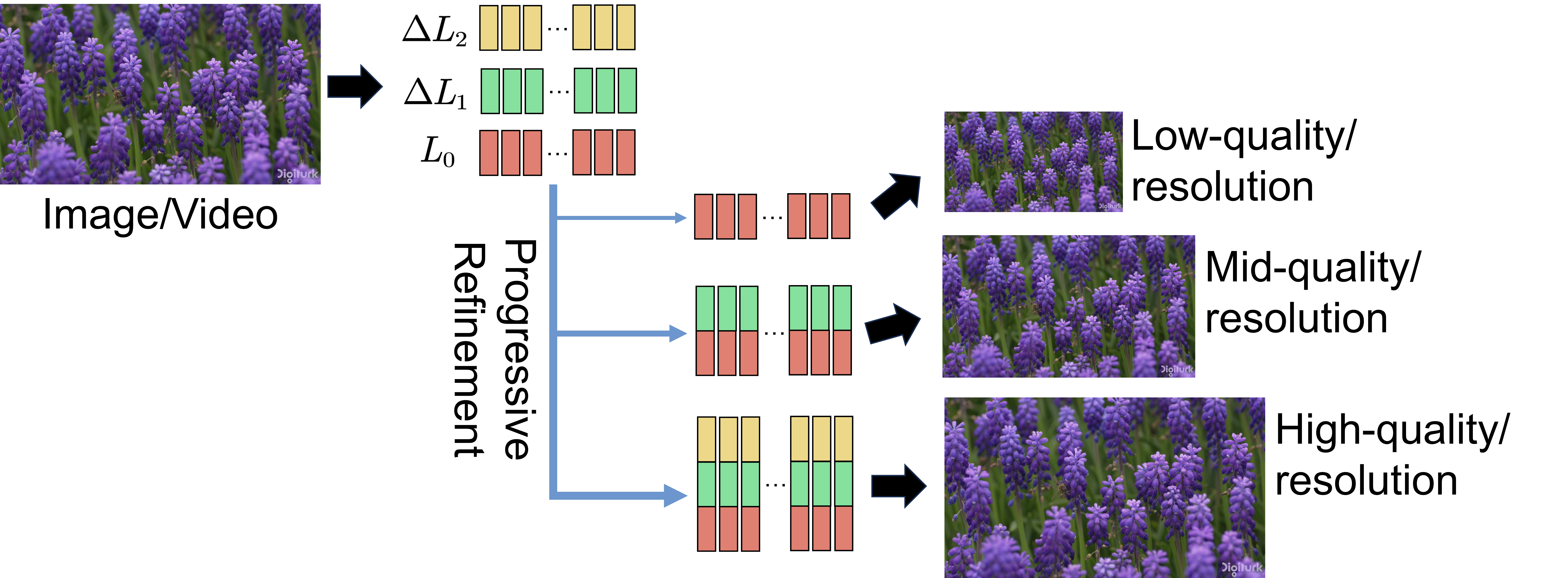}
  \caption{The architecture of \name, shown with three layers as an example. The base layer $L_0$ provides the coarse representation, while the enhancement layers $\Delta L_1$ and $\Delta L_2$ progressively add additional 2D Gaussian splats. By incrementally incorporating these enhancement layers, the video quality is refined progressively, yielding higher-quality/resolution reconstructions.
 }
 \label{fig:Insight}
\end{figure}

\begin{figure*}[t]
    \centering
    \subfloat[$\boldsymbol{\mathcal{G}}^{0}$]{\includegraphics[width=0.245\textwidth]{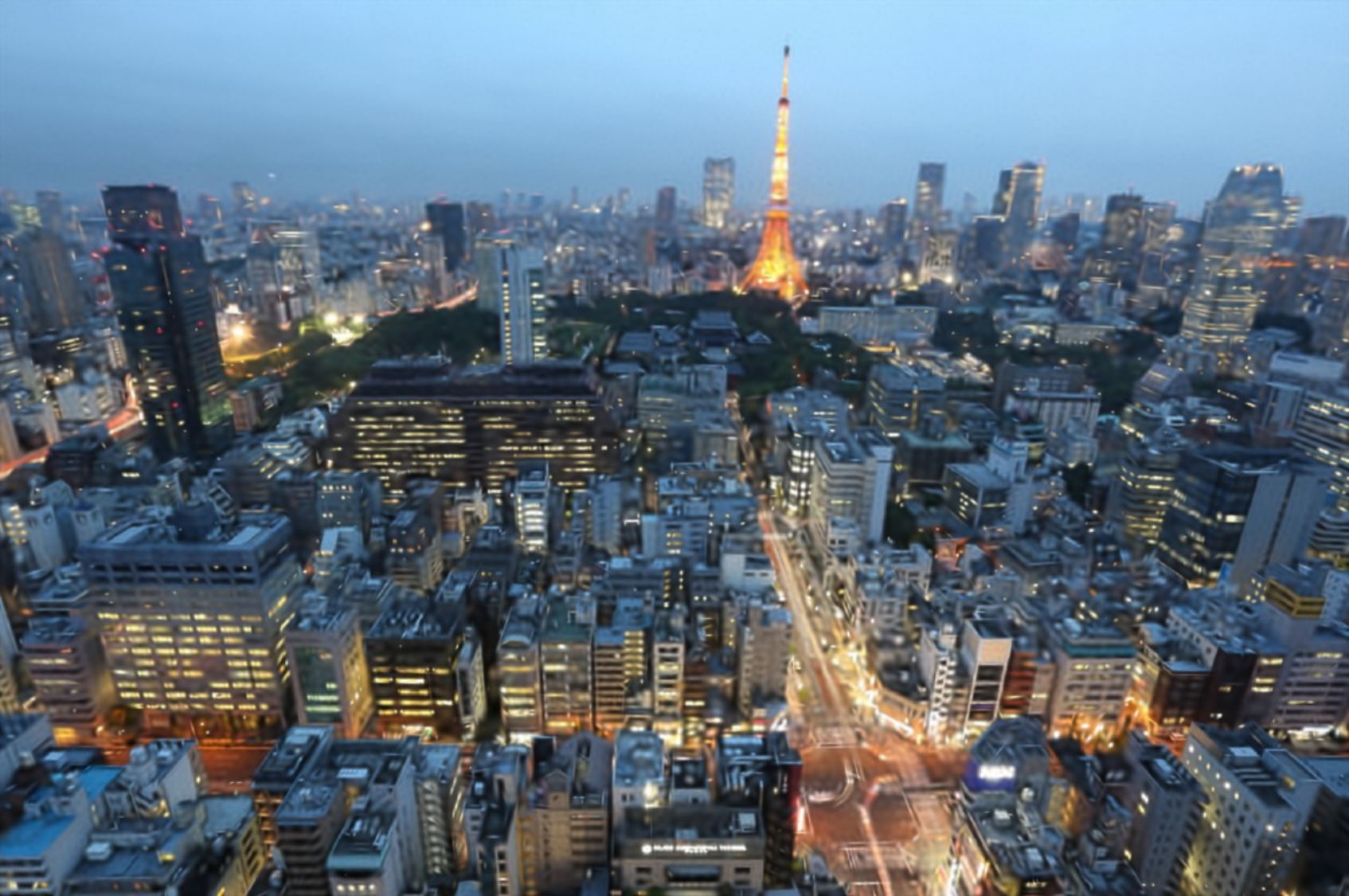}}\hspace{1px}
    \subfloat[$\Delta\boldsymbol{\mathcal{G}}^{1}$]{\includegraphics[width=0.245\textwidth]{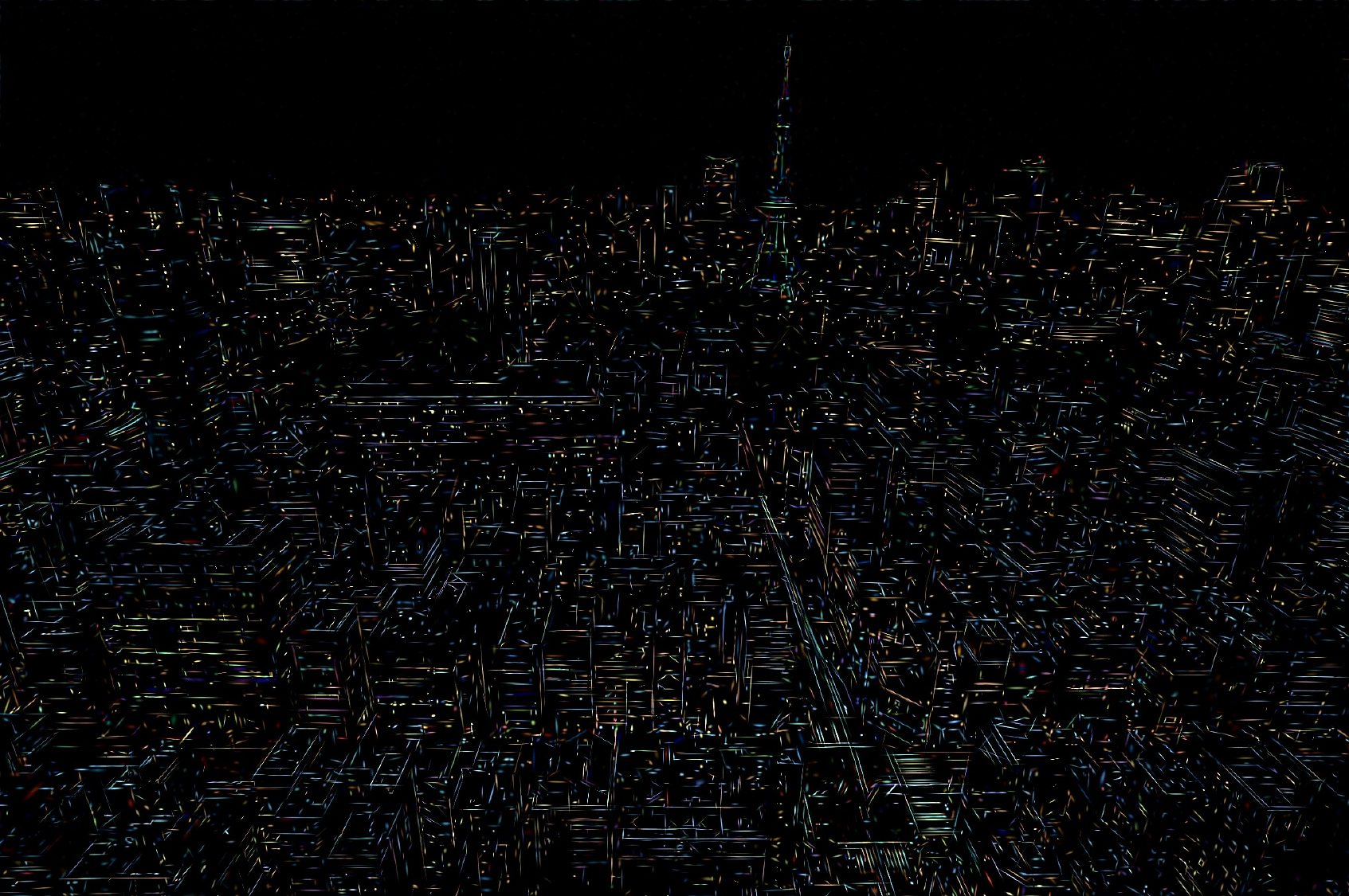}}\hspace{1px}
    \subfloat[$\Delta\boldsymbol{\mathcal{G}}^{2}$]{\includegraphics[width=0.245\textwidth]{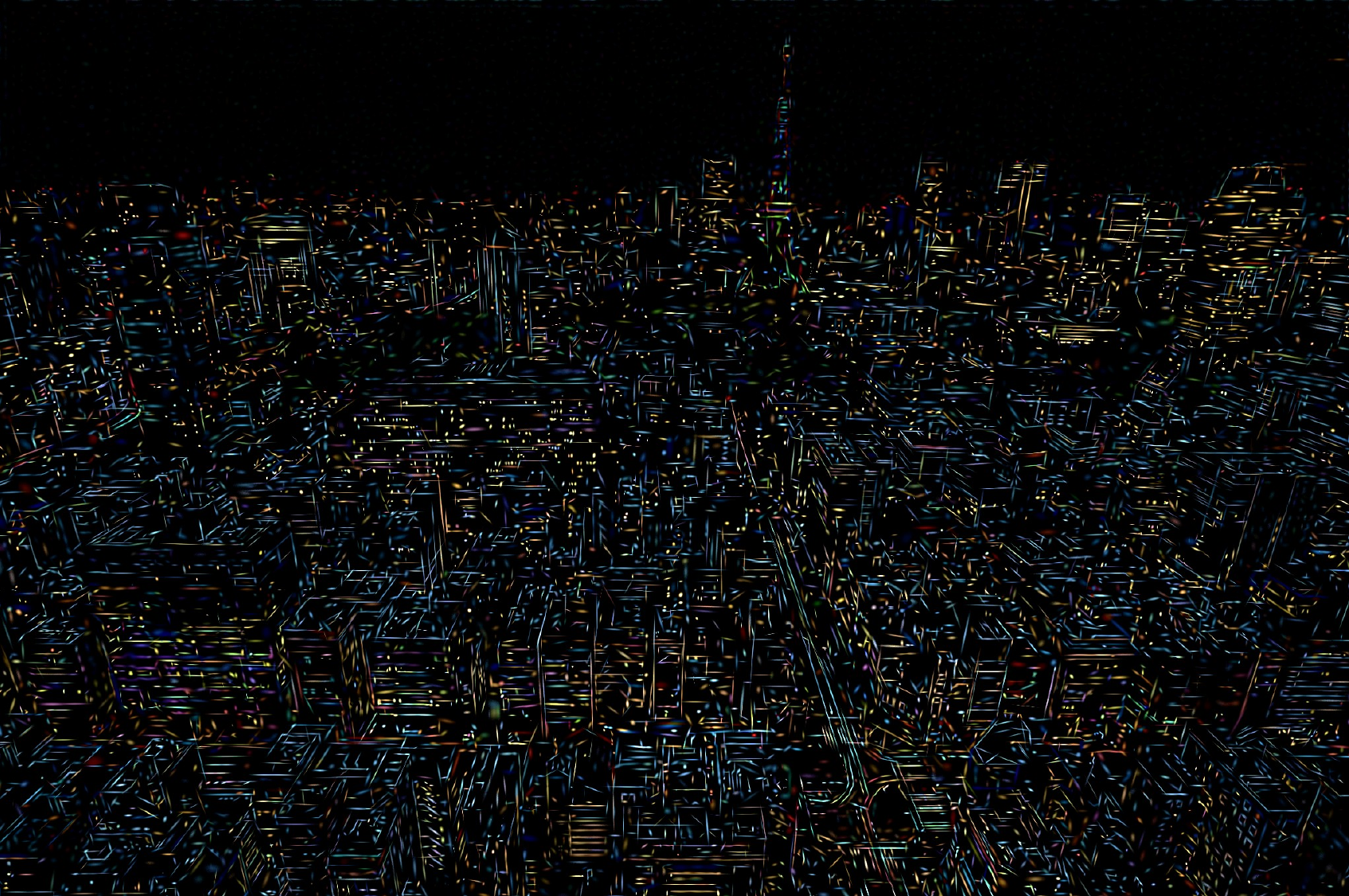}}\hspace{1px}
    \subfloat[$\boldsymbol{\mathcal{G}}^{2}$]{\includegraphics[width=0.245\textwidth]{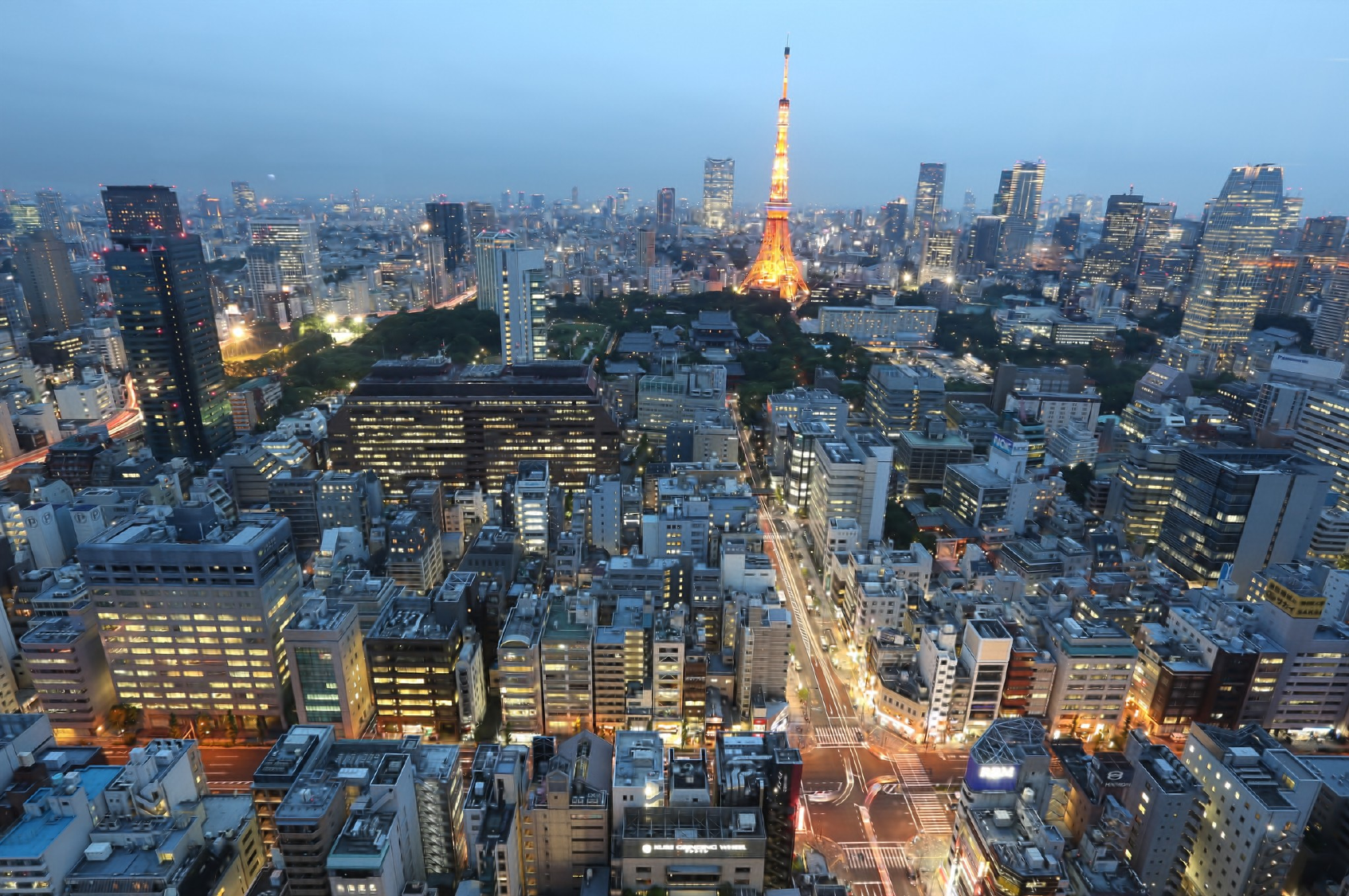}}
    \caption{
    Progressive procedure of layered Gaussians.  $\Delta\boldsymbol{\mathcal{G}}^{1}$ and $\Delta\boldsymbol{\mathcal{G}}^{2}$  capture higher-frequency details and can be iteratively added to $\boldsymbol{\mathcal{G}}^{0}$, yielding progressively refined $\boldsymbol{\mathcal{G}}^{2}$.}
    \label{fig:progressive}
\end{figure*}

We address the challenges of training a progressive, scalable version of GSCV in this paper. 
We propose Progressive Gaussian Splat Video Coding (\name), a progressive scalable video codec that uses 2D Gaussian splats as primitives.  
\name{} applies to scalable image coding as well, treating the image as a single-frame video.
~\Cref{fig:Insight} shows the overview of {\name}, where an input video is represented by multiple layers of Gaussian splats.  Just as a typical layered progressive codec, decoding only the base layer reconstructs a low-fidelity version of the input, and decoding additional enhancement layers increases the fidelity.  
~\Cref{fig:progressive} visualizes the rendering of the splats in each layer independently.  
To avoid cross-layer optimization conflicts and ensure smooth progressivity, {\name} uses a joint training strategy. 
In each iteration, it simultaneously optimizes two fidelity levels: one using all splats up to the highest layer, and another using only splats up to some layer $i$.  During training, we cycle the target layer $i$ from the base layer to the second-highest layer. 
This joint optimization allows splats' optimization trajectories to be aligned across layers while each layer's splats are still trained to meet the layer's objectives.
Furthermore, by cycling through $i$, the gradient field remains relatively stable during transitions between optimization objectives, ensuring that both final and intermediate reconstructions converge to acceptable quality.

The main contributions of this paper are summarized as follows:
\begin{itemize}[leftmargin=*]
    \item We introduce a scalable progressive representation framework for both image and video using 2DGS. 
    \item We demonstrate the effectiveness of 
    a joint training strategy for achieving progressivity in 2DGS.   
    \item We evaluated {\name} against existing training methods for 3DGS and 2DGS, showing that {\name} can achieve up to 1.9~dB improvement in PSNR (over UVG video dataset) and 2.6~dB improvement in PSNR (over DIV-HR image dataset), compared to methods that performs sequential layer-wise training.
\end{itemize}

\section{Related Work} \label{sec:related}
\subsection{Scalable Coding for Image and Video}
Scalable coding encodes visual content (e.g., images or videos) into a single bitstream that supports decoding of multiple versions, enabling adaptive delivery across heterogeneous devices and varying network conditions.
In scalable image coding, classical methods such as JPEG2000~\cite{DBLP:books/daglib/0007442} employ wavelet-based hierarchies to support resolution and quality scalability within a single bitstream.
Building upon similar principles, scalable video coding introduces temporal dependencies and inter-layer prediction across frames. 
Early standards such as MPEG-2~\cite{ISO13818-2} introduced layered spatial and quality scalability using base and enhancement layers.
MPEG-4 Visual~\cite{ISO14496-2} extended quality scalability through Fine Granularity Scalability (FGS), enabling progressive truncation of enhancement-layer bitstreams.
The scalable extension of H.264/AVC~\cite{DBLP:journals/tcsv/SchwarzMW07} later formalized a structured and coding-efficient multi-layer framework supporting spatial, temporal, and quality scalability, which was further streamlined and extended to bit-depth and color-gamut scalability in HEVC-SHVC~\cite{DBLP:journals/tcsv/BoyceYCR16}.
Yet despite these improvements, they have seen limited deployment in practice, in part due to the codec complexity.

With the rise of deep learning, learning-based methods have emerged that simplify the pipeline through end-to-end optimization.
These approaches generally follow two paradigms.
One represents images or videos in the latent space, where hierarchical decoders sample at different scales to reconstruct at different quality and enable progressive decoding~\cite{DBLP:conf/wacv/PrestaTFGC25, DBLP:conf/aaai/LuD0M24,DBLP:conf/mm/WuQHLLYLY24}.
The other encodes an image or video directly into network parameters, where the base and enhancement subnetworks correspond to different quality levels, so that quality improves as more parameters are used~\cite{DBLP:journals/tomccap/ZhangZT24, DBLP:conf/mmm/CaoZS24}.
However, both paradigms rely on implicit representations that hinder manipulation and the functionality of classical scalable codecs, while also incurring a high decoding cost that restricts practical deployment.
In contrast, our \name~ leverages a more explicit primitive, 2D Gaussian splat, to achieve progressive scalability via the addition or removal of splats.

\subsection{Gaussian Splatting for Image and Video}
Gaussian splatting represents visual content with explicit 2D or 3D Gaussian primitives, where fidelity can be easily controlled by the number and parameters of Gaussian splats.
For images, GaussianImage~\cite{DBLP:conf/eccv/ZhangGXHWQLGZ24} uses 2D Gaussian splats for image representation and compression, while LIG~\cite{DBLP:conf/aaai/ZhuLCZJWY25} extends it for high-quality large-image representation.
For videos, existing works can be categorized into 3D and 2D. 
For 3D Gaussian Video, VGR~\cite{DBLP:conf/nips/Sun0MLC024} treats a video as a sequence of static 3D scenes, reconstructing each frame independently with 3DGS. 
By contrast, VeGaS~\cite{DBLP:journals/corr/abs-2411-11024} and GaussianVideo~\cite{DBLP:journals/corr/abs-2501-04782} view the entire sequence as a continuous spatio-temporal volume, where splats evolve smoothly over time. 
These 3D approaches excel at video representation and editing but are not tailored for compression or scalable coding.
In contrast, 2D Gaussian Video encodes each frame with lightweight 2DGS~\cite{DBLP:conf/eccv/ZhangGXHWQLGZ24} primitives, enabling video representation and compression~\cite{DBLP:conf/cvpr/LeeCL25,DBLP:conf/nossdav/WangSO25}.
However, extending 2D Gaussian representation to scalable coding is not trivial, as discussed in~\Cref{Sec:Introduction}.
In this work, we extend 2DGS by reorganizing Gaussian splats into a multi-layer structure, enabling progressive, scalable image and video coding.

\subsection{Progressive Gaussian Splatting}
Progressive Gaussian Splatting reconstructs content in a coarse-to-fine manner, where a subset of splats yields a low-fidelity preview, and additional splats progressively refine details. 
The main advantage of progressive Gaussian splatting is that it supports partial decoding for fast low-fidelity previews and progressive refinement toward full quality, thereby providing inherent scalability and adaptability to varying bandwidth and computational resources.
Progressive Gaussian splatting was first explored in FreGS~\cite{DBLP:conf/cvpr/ZhangZXLX24} and later extended for rendering (PRoGS~\cite{DBLP:conf/wacv/Zoomers0MVPM25}), streaming (LapisGS~\cite{DBLP:conf/3dim/ShiMGO25} and its dynamic extension~\cite{sun2025tsla}), and compression (PCGS~\cite{DBLP:journals/corr/abs-2503-08511}), in the context of 3DGS.
In contrast, progressive coding for lightweight 2DGS, better suited to image and video transmission, has not been extensively explored.
The first attempt is LIG~\cite{DBLP:conf/aaai/ZhuLCZJWY25}, which reconstructs images by a two-layer Gaussian structure, by separately training a skeleton layer and a residual layer. 
While effective for large images, LIG prioritizes final reconstruction quality over scalable coding: achieving good visual quality requires many splats for high-frequency details, so under limited bandwidth, where only a small subset can be transmitted, performance degrades significantly.
Moreover, its sequential layer-wise training strategy leads to conflicting optimization objectives across layers as 
mentioned in~\Cref{Sec:Introduction}.

\section{\name} \label{sec:methodology}
\subsection{Problem Formulation \label{sec:formulation}}
We first introduce the problem formulation of our progressive, scalable Gaussian representation for both image and video.
We distinguish two terms: ``layer'' refers to the structural components in the layered representation, while ``levels'' refers to the reconstructed image or video formed by accumulating layers.
Let $\boldsymbol{\mathcal{F}}=\{f_1,f_2,\dots,f_T\}$ denote the input sequence, where $T$ is the number of frames (for images, $T=1$).
Specifically, given $L$ layers, our goal is to encode$\boldsymbol{\mathcal{F}}$ into a hierarchical representation:
$
\widehat{\boldsymbol{\mathcal{F}}}
= \widehat{\boldsymbol{\mathcal{F}}} ^{0}\cup\left(\bigcup_{\ell=1}^L \Delta \widehat{\boldsymbol{\mathcal{F}}} ^{\ell}\right),
$
where $\widehat{\boldsymbol{\mathcal{F}}} ^{0}$ serves as the base layer as well as the base level, and $\Delta\widehat{\boldsymbol{\mathcal{F}}} ^{\ell}$ ($\ell=1,\dots, L$) are the enhancement layers.
For any $\ell \in \{1, 2,\dots,L\}$, the reconstruction at level $\ell$ is obtained by progressively adding the first $\ell$ enhancement layers to the base level:
\begin{equation}
\widehat{\boldsymbol{\mathcal{F}}} ^{\ell} 
= \widehat{\boldsymbol{\mathcal{F}}} ^{0} + \sum_{i=1}^{\ell}\Delta\widehat{\boldsymbol{\mathcal{F}}} ^{i}.
\end{equation}
In ~\name, $\widehat{\boldsymbol{\mathcal{F}}}$ is represented by a set of 2D Gaussian primitives $\boldsymbol{\mathcal{G}} = \{\boldsymbol{\mathcal{G}}_{1}, \boldsymbol{\mathcal{G}}_{2}, \dots, \boldsymbol{\mathcal{G}}_{T}\}$ to reconstruct $\boldsymbol{\mathcal{F}}$ while preserving scalability. Note that ~\name~ supports both scalable video ($T>1$) and image ($T=1$). 


\subsection{2D Gaussian Splats \label{sec:2dGaussian}}
We adopt 2D Gaussian splats~\cite{DBLP:conf/eccv/ZhangGXHWQLGZ24} as the fundamental primitives in our framework.
Each Gaussian splat $G$ is parameterized by a tuple $G = (\boldsymbol{\mu}, \boldsymbol{\Sigma}, \mathbf{c}')$, where $\boldsymbol{\mu}\in\mathbb{R}^2$ denotes its center in the image plane, $\boldsymbol{\Sigma}\in\mathbb{R}^{2\times 2}$ is the covariance matrix that controls scale and orientation, and $\mathbf{c}'\in\mathbb{R}^3$ is a learned representation for the associated RGB color and opacity.
To ensure that $\boldsymbol{\Sigma}$ is positive semi-definite, Cholesky decomposition~\cite{DBLP:reference/opt/Haddad09} is applied, representing it as $\boldsymbol{\Sigma}=\boldsymbol{L}\boldsymbol{L}^\top$, where $\boldsymbol{L}$ is a lower-triangular matrix parameterized by the Cholesky vector $\boldsymbol{\ell}=(\ell_{1},\ell_{2},\ell_{3})$.
In total, each 2D Gaussian splat requires only eight parameters, yielding a compact and efficient representation.
For rendering, the color $\boldsymbol{C}_{i}$ of the $i$-th pixel is computed as the weighted accumulation of all splats covering that pixel:  
\begin{equation}
\label{equ:rendering}
    \boldsymbol{C}_{i} = \sum_{n \in \mathcal{N}} \boldsymbol{c}'_{n}   
    \exp\!\left(-\tfrac{1}{2}\boldsymbol{d}_{n}^\top \boldsymbol{\Sigma}_{n}^{-1} \boldsymbol{d}_{n}\right),
\end{equation}
where $\mathcal{N}$ is the set of splats overlapping pixel $i$, and $\boldsymbol{d}_{n}$ denotes the distance between the pixel center and the center of the $n$-th Gaussian. 

In \name, we further follow the rendering scheme in GSVC~\cite{DBLP:conf/nossdav/WangSO25}, where an additional weight vector $\boldsymbol{w}$ is introduced to track the contribution of Gaussian splats.
This allows splats to be dynamically added or removed based on their contribution, enabling a trade-off between quality and size as well as enhancement of highly dynamic content.
Accordingly, the rendering formula in~\Cref{equ:rendering} is written as:
\begin{equation}
\label{equ:rendering_rgbW}
    \boldsymbol{C}_{i} = \sum_{n \in \mathcal{N}} w_n    \boldsymbol{c}'_{n}   
    \exp\!\left(-\tfrac{1}{2}\boldsymbol{d}_{n}^\top \boldsymbol{\Sigma}_{n}^{-1} \boldsymbol{d}_{n}\right).
\end{equation}

Moreover, 2DGS provides advantages for scalable coding.
First, the representation itself is resolution-agnostic: a fixed set of splats can be rendered at arbitrary resolutions without retraining (though quality may degrade if the rendering resolution substantially differs from the training resolution).
Second, its explicit and modular structure supports scalability by allowing a subset of splats to be decoded independently.  

\subsection{Progressive Gaussian Splatting \label{sec:progressive_coding}}

For progressive scalability, we represent each input (either a video or an image) as a layered collection of Gaussian sets.
For the base level representation $\widehat{\boldsymbol{\mathcal{F}}}^0$, each frame $\widehat{\boldsymbol{f}}^{0}_{t}$ is associated with a Gaussian set
$\boldsymbol{\mathcal{G}}^0_t = \{\boldsymbol{G}_{t,1}, \boldsymbol{G}_{t,2}, \dots, \boldsymbol{G}_{t,N^0}\}$,
where $N^0$ is the number of splats in the base layer for each frame.
Enhancement layers $\Delta\widehat{\boldsymbol{\mathcal{F}} }^{\ell}$ ($\ell=1,\dots,L$) are defined similarly, with each frame represented by
$\Delta\boldsymbol{\mathcal{G}}^\ell_t = \{\boldsymbol{G}_{t,1}, \boldsymbol{G}_{t,2}, \dots, \boldsymbol{G}_{t,N^\ell}\}$,
where $N^\ell$ is the number of splats in the $\ell$-th layer for each frame. 
The layered Gaussian representation of a frame at level $\ell$ is then obtained by the union of the base and the first $\ell$ enhancement Gaussian sets:
\begin{equation}
    \boldsymbol{\mathcal{G}}^\ell_t 
    = \boldsymbol{\mathcal{G}}^0_t  \cup  \left(\bigcup_{i=1}^{\ell} \Delta\boldsymbol{\mathcal{G}}^i_t\right).
\end{equation}
Correspondingly, $\widehat{\boldsymbol{\mathcal{F}} }^{\ell}$ is represented as
$    \boldsymbol{\mathcal{G}}^\ell = \{\boldsymbol{\mathcal{G}}^\ell_1, \dots, \boldsymbol{\mathcal{G}}^\ell_T\}$.
Its reconstruction is obtained by rendering the Gaussian sets with the differentiable rasterizer $\mathcal{R}(\cdot)$:
\begin{equation}
    \label{equ:render}
    \widehat{\boldsymbol{\mathcal{F}}}^{  \ell}
    = \mathcal{R}\!\left(\boldsymbol{\mathcal{G}}^\ell\right),
    \qquad
    \widehat{\boldsymbol{f}}^{  \ell}_t = \mathcal{R}\!\left(\boldsymbol{\mathcal{G}}^\ell_t\right).
\end{equation}
The operator $\mathcal{R}(\cdot)$ follows the pixel-wise accumulation rule in~\Cref{equ:rendering_rgbW}, applied to the unioned Gaussian set $\boldsymbol{\mathcal{G}}^\ell_t$.

This layered construction enables progressive decoding: the base set $\boldsymbol{\mathcal{G}}^0_t$ provides a coarse but complete representation, while each $\Delta\boldsymbol{\mathcal{G}}^i_t$ adds finer details. Thus, increasing $\ell$ monotonically improves quality without modifying previously transmitted splats. As illustrated in~\Cref{fig:progressive}, $\boldsymbol{\mathcal{G}}^0$ yields the coarse reconstruction, $\Delta \boldsymbol{\mathcal{G}}^1$ and $\Delta \boldsymbol{\mathcal{G}}^2$ successively refine details, and their union $\boldsymbol{\mathcal{G}}^2 = \boldsymbol{\mathcal{G}}^0 \cup \Delta \boldsymbol{\mathcal{G}}^1 \cup \Delta \boldsymbol{\mathcal{G}}^2$ produces the higher-quality result.

\subsection{Joint Training \label{sec:joint_training}}
A baseline approach to optimizing the layered splats is sequential layer-wise training. 
Specifically, for level $\ell$, the reconstruction is written as
\begin{equation}
    \widehat{\boldsymbol{\mathcal{F}}}^{\ell} = \widehat{\boldsymbol{\mathcal{F}}}^{\ell-1} + \mathcal{R}\!\left(\Delta \boldsymbol{\mathcal{G}}^\ell\right),
\end{equation}
where $\widehat{\boldsymbol{\mathcal{F}}}^{\ell-1}$ is the reconstruction from the previous layers and $\mathcal{R}(\Delta \boldsymbol{\mathcal{G}}_\ell)$ denotes the rendering of the newly introduced splats in layer $\ell$. 
The training process proceeds by first optimizing the base set $\boldsymbol{\mathcal{G}}^0$, then freezing its parameters, and subsequently training the enhancement sets $\Delta \boldsymbol{\mathcal{G}}^1, \Delta \boldsymbol{\mathcal{G}}^2, \dots, \Delta \boldsymbol{\mathcal{G}}^L$ sequentially. 
This strategy is analogous to LIG~\cite{DBLP:conf/aaai/ZhuLCZJWY25}, which represents large images with two Gaussian layers trained sequentially, freezing the first as the foundation for the second.

However, such a layer-wise method leads to several issues.
First, since different layers pursue different optimization objectives, their optimization follows distinct trajectories. 
When a higher layer is trained on top of already converged lower layers, the optimization must adapt to a different objective landscape, 
hindering convergence.
As shown in~\Cref{fig:sequentially_layer-wise}, switching the target layer at the 50k and 100k iterations causes sharp fluctuations in the gradients, indicating that the model struggles to adapt to a change in the objective.  
Compared to single-layer GSVC optimization in~\Cref{fig:single-layer}, layer-wise training converges to a higher loss value, indicating that the model 
is trapped in a suboptimal local minima.

To optimize layered splats effectively, the two key challenges, cross-layer optimization conflict and progressive stability, should be considered. 

\begin{figure}[t]
\centering
\begin{minipage}[b]{0.49\textwidth}\centering
  \subfloat[Uniformly Random Level Selection]{
    \includegraphics[width=0.49\textwidth]{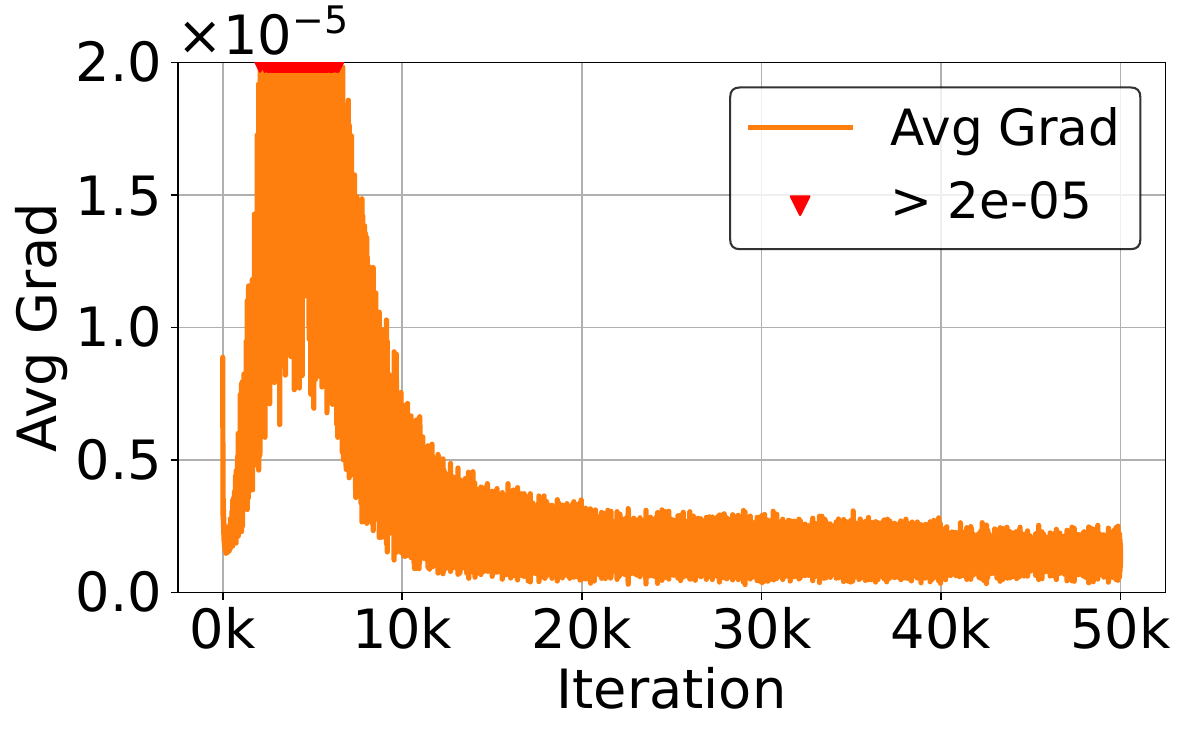}
    \includegraphics[width=0.49\textwidth]{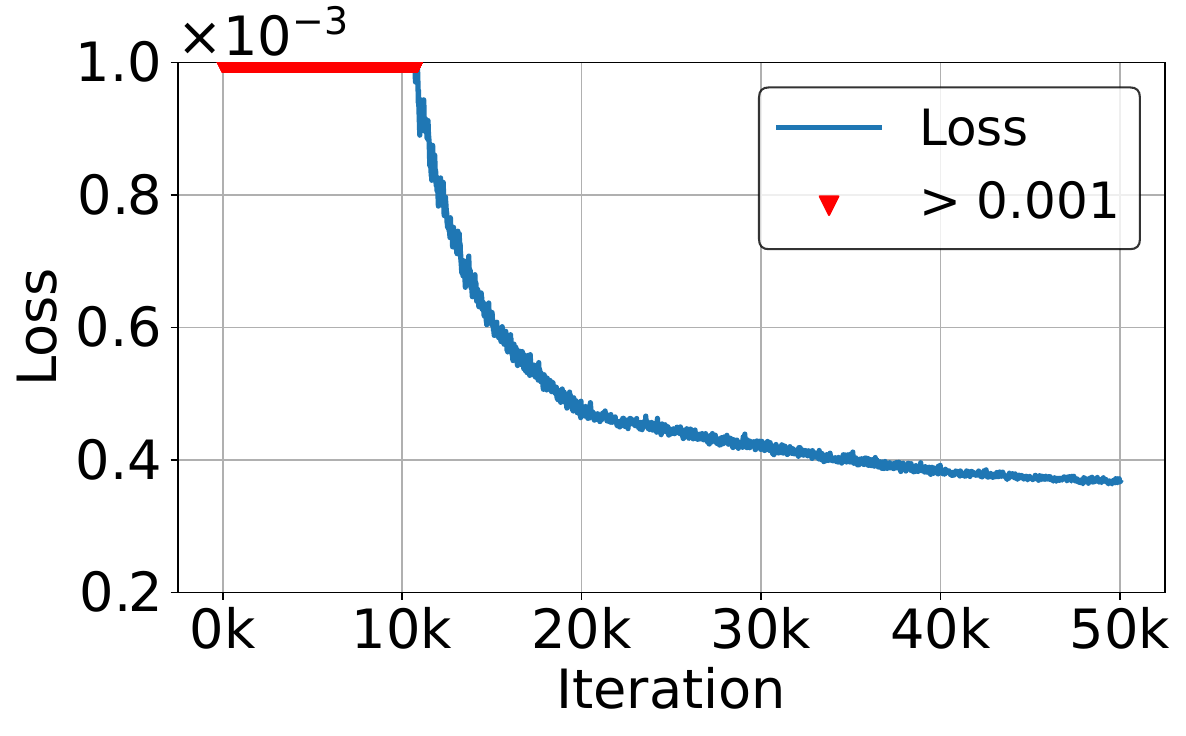}
    \label{fig:randomly_Joint}
  }\hfill
\end{minipage}
\begin{minipage}[b]{0.49\textwidth}\centering
  \subfloat[Cyclic Level Selection]{
    \includegraphics[width=0.49\textwidth]{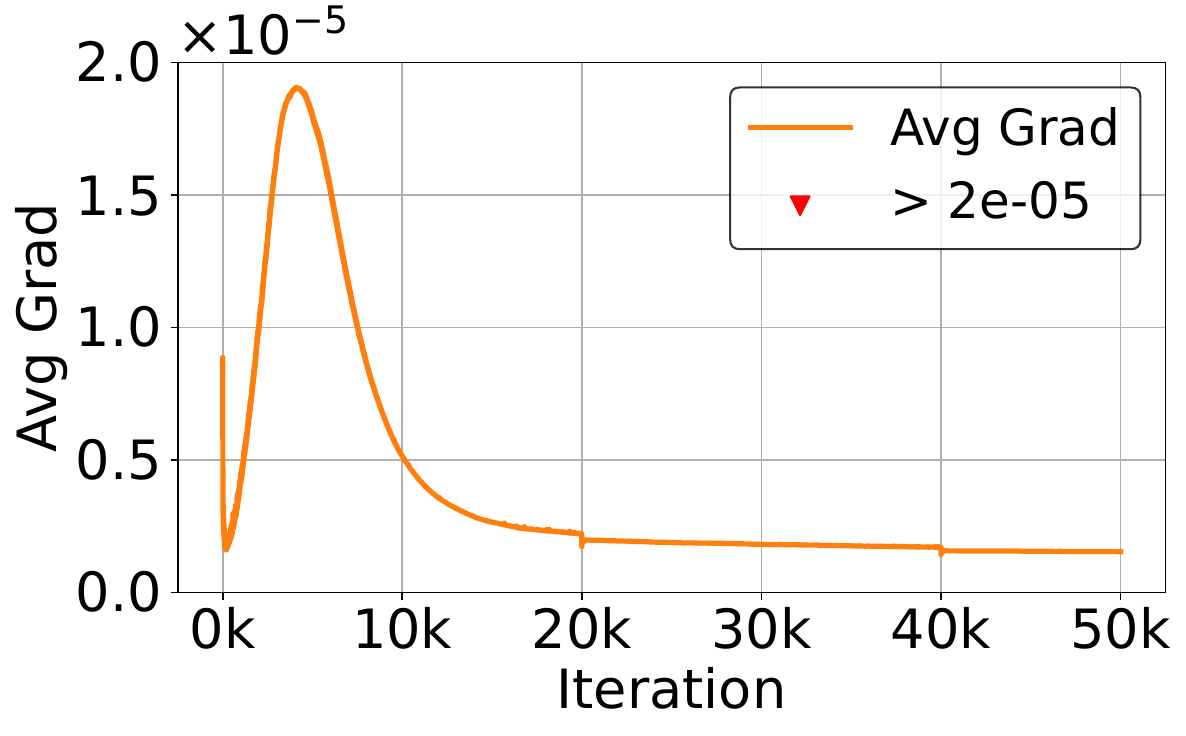}
    \includegraphics[width=0.49\textwidth]{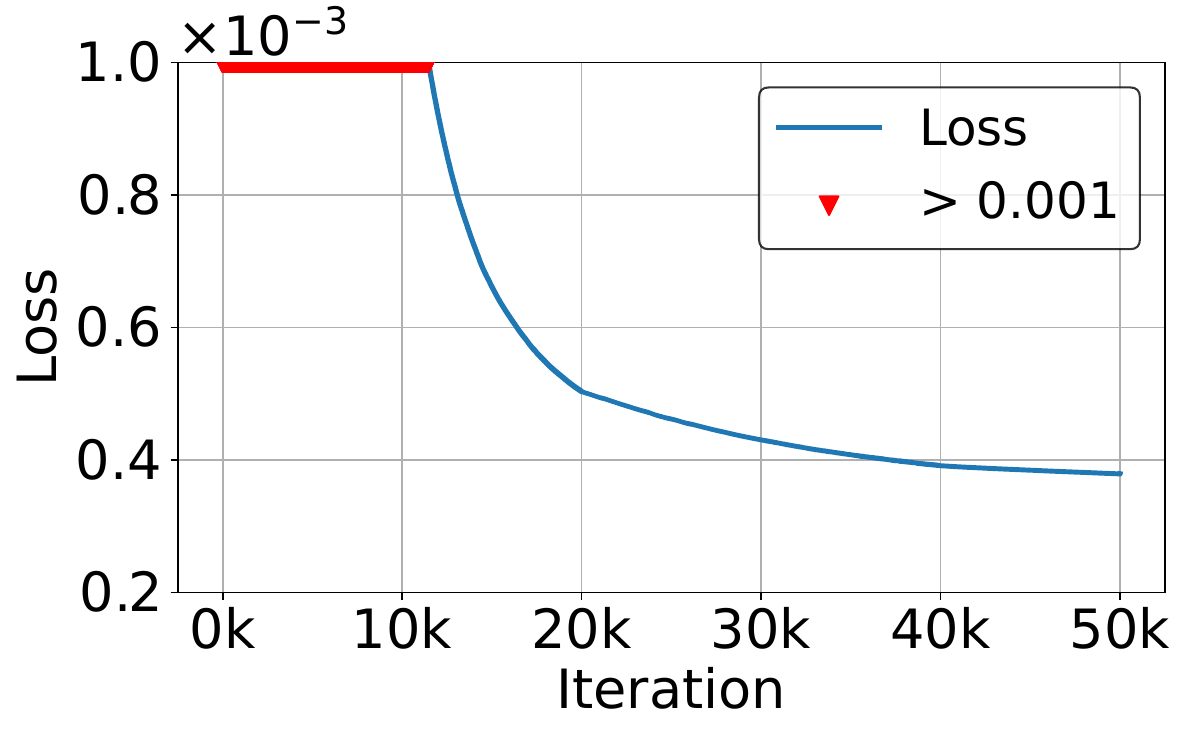}
    \label{fig:cyclically_Joint}
  }\hfill
\end{minipage}
\vspace{-.2cm}
\caption{Comparison of gradient and loss over iterations on randomly and cyclically joint training strategies: jointly optimizes layers  (a) randomly and (b) in a cyclic order.}
\label{fig:insight-comparision-Joint}
\vspace{-.2cm}
\end{figure}

\textbf{Cross-Layer Optimization Conflicts}. {\name} requires a coherent multi-layer solution, yet this is challenging since different layers pursue distinct objectives: The base layer captures coarse structures, while enhancement layers refine the details.
To improve the compatibility of the optimization objectives across the layers, we jointly train the layers belonging to multiple levels, rather than training them separately, allowing splats in different layers to be optimized along aligned trajectories, using a joint loss function. This process encourages splats in each layer to be compatible with each other from the outset.
Given the $t$-th frame $\boldsymbol{f}_t$ as an example, at each iteration, we supervise both the full reconstruction $\widehat{\boldsymbol{f}}_{t}^{L}$ and intermediate reconstructions $\widehat{\boldsymbol{f}}_{t}^{\ell}$. 
Formally, the loss function is defined as the sum of $\mathcal{L}_2$ losses across selected levels:
\begin{equation}
    \mathcal{L}_{t} =  \sum_{\ell \in \mathcal{S}}\mathcal{L}_2(\widehat{\boldsymbol{f}}_{t}^{\ell},\boldsymbol{f}_{t})= \sum_{\ell \in \mathcal{S}} 
    \frac{1}{HW} \sum_{i=1}^{HW} 
    \big\| \widehat{f}^{  \ell}_{t,i} - f_{t,i} \big\|^2,
\end{equation}
where $\widehat{f}^{\ell}_{t, i}$ denotes the reconstructed $i$-th pixel in $f_t$ at level $\ell$, $H$ and $W$ are the spatial dimensions, and $\mathcal{S}$ is the set of layers optimized jointly in the current iteration. 
In practice, $\mathcal{S}$ contains both the final level and one intermediate level $\ell \in \{1,2,\cdots,L-1\}$. 
Each level is supervised against the ground truth at its corresponding resolution. 
This setting ensures that lower layers focus on coarse structures while higher layers refine fine details, and preventing aliasing at lower levels~\cite{DBLP:conf/aaai/ZhuLCZJWY25}. 
The loss can be written as:
\begin{equation}
\label{equ:L}
    \mathcal{L}_t(\ell) 
    = \mathcal{L}_2\!\big(\widehat{\boldsymbol{f}}_{t}^{  L}, \boldsymbol{f}_{t}^{  L}\big) 
    + \mathcal{L}_2\!\big(\widehat{\boldsymbol{f}}_{t}^{  \ell}, \boldsymbol{f}_{t}^{  \ell}\big),
\end{equation}
where $\boldsymbol{f}_{t}^{  \ell} \in \mathbb{R}^{H_\ell \times W_\ell}$ is the corresponding ground-truth frame downsampled to resolution $(H_\ell,W_\ell)$.

\textbf{Progressive Stability}. 
For progressive scalability, intermediate reconstruction quality must remain high, ideally comparable to the non-layered version with the same number of splats.
To this end, we adopt cyclic joint training, where the intermediate level is chosen in an incremental cyclic order, 
rather than uniformly and randomly during iterative training. 

%
A naive uniformly random level selection can lead to the same level being chosen consecutively, leading to overfitting, or a level not being chosen for some number of iterations, leading to misalignment with the optimization process for reconstruction of the highest level.
As shown in~\Cref{fig:randomly_Joint}, random joint training causes unstable gradients and irregular loss updates. By contrast, cyclic selection enforces a fixed periodic switching of intermediate levels, avoiding idle intervals and consecutive repetitions, and thus yields smoother optimization trajectories with more stable gradients and losses, as shown in~\Cref{fig:cyclically_Joint}.

Finally, the loss at iteration $k$ for the $t$-th frame $\boldsymbol{f}_{t}\in\boldsymbol{\mathcal{F}}$ is defined as:
\begin{equation}
\label{equ:k-L}
    \mathcal{L}_t(\ell_k) 
    = \mathcal{L}_2\!\big(\widehat{\boldsymbol{f}}_{t}^{L}, \boldsymbol{f}_{t}^{  L}\big) 
    + \mathcal{L}_2\!\big(\widehat{\boldsymbol{f}}_{t}^{\ell_k}, \boldsymbol{f}_{t}^{  \ell_k}\big),
\end{equation}
where $\ell_k \equiv k  \big(\mathrm{mod} (L-1)\big)$ denotes the intermediate level selected in cyclic order.

\subsection{Put Everything Together\label{sec:put_together}}

\begin{algorithm*}[t]
\caption{\name~ Training Pipeline}
\label{alg:training}
\begin{algorithmic}[1]
\REQUIRE Input sequence $\boldsymbol{\mathcal{F}}=\{\boldsymbol{f}_1, \boldsymbol{f}_2, \dots, \boldsymbol{f}_T\}$, number of layers $L$, number of splats ($N$, $N_{aug}$, $N_{prune}$), GSP configuration ($Z_{GSP}, K_{GSP}$), target resolutions $\{(H_\ell,W_\ell)\}_{\ell=0}^L$.
\STATE  Obtain $\{\boldsymbol{\mathcal{F}}^\ell\}_{\ell=1}^L$ by downsampling $\boldsymbol{\mathcal{F}}$ given $\{(H_\ell,W_\ell)\}_{\ell=0}^L$. 
\IF{$T>1$}
\STATE Select I-frames using DKS.
\ENDIF
\FOR{$t=1$ to $T$}
    \IF{$T > 1$ (video) and $t$ is an I-frame \textbf{or} $T=1$ (image)}
        \STATE Initialize $\boldsymbol{\mathcal{G}}_t = \{\boldsymbol{\mathcal{G}}_t^0, \Delta \boldsymbol{\mathcal{G}}_t^1, \dots, \Delta \boldsymbol{\mathcal{G}}_t^L\}$ with $\boldsymbol{N}+\boldsymbol{N}_{aug}$ random Gaussian splats (uniformly spread over all layers).
    \ELSE
        \STATE Initialize $\boldsymbol{\mathcal{G}}_t = \{\boldsymbol{\mathcal{G}}_t^0, \Delta \boldsymbol{\mathcal{G}}_t^1, \dots, \Delta \boldsymbol{\mathcal{G}}_t^L\}$ with $\boldsymbol{\mathcal{G}}_t^0 = \boldsymbol{\mathcal{G}}_{t-1}^0$ and $\Delta \boldsymbol{\mathcal{G}}_t^i = \Delta \boldsymbol{\mathcal{G}}_{t-1}^i$ for $i = 1 \ldots L$.
        \STATE Inject $N_{aug}$ random Gaussian splats (uniformly spread over all layers).
    \ENDIF
    \WHILE{not converged, iteration $k=0,1,2,\dots$}
        \STATE Cyclically select $\ell_k$: $\ell_k \equiv k  \big(\mathrm{mod} (L-1)\big)$.
        \STATE Progressively render $\widehat{\boldsymbol{f}}_{t}^{\ell_k}$ and $\widehat{\boldsymbol{f}}_{t}^L$ by~\Cref{equ:render}.
        \STATE Calculate joint loss by~\Cref{equ:k-L} and update $\boldsymbol{\mathcal{G}}_t$.
        \STATE Remove $\boldsymbol{N}_{\mathrm{prune}} \cdot \frac{Z_{\mathrm{GSP}}}{K_{\mathrm{GSP}}}$ splats with minimal contribution according to $\boldsymbol{w}$ in~\Cref{equ:rendering_rgbW} every $Z_\mathrm{GSP}$ iteration in the first $K_\mathrm{GSP}$ iteration.
\ENDWHILE
\ENDFOR
\ENSURE Optimized layered splats $\boldsymbol{\mathcal{G}} = \{\boldsymbol{\mathcal{G}}^0, \Delta \boldsymbol{\mathcal{G}}^1, \dots, \Delta \boldsymbol{\mathcal{G}}^L\}$.
\end{algorithmic}
\end{algorithm*}

In \name, to achieve efficient 2DGS representation, we borrow the core paradigm of GSVC~\cite{DBLP:conf/nossdav/WangSO25}.
For videos, frame prediction exploits temporal redundancy by predicting each P-frame from its predecessor;
Gaussian Splat Pruning (GSP) removes splats with low contribution to balance quality and bitrate;
Gaussian Splat Augmentation (GSA) introduces new splats to capture highly dynamic regions;
and Dynamic Key-frame Selection (DKS) detects scene transitions to insert I-frames adaptively.
For the image case, we treat it as a single I-frame video without temporal redundancy.

The complete training procedure for resolution scalability of {\name} is summarized in Algorithm~\ref{alg:training}.

We first construct a multi-resolution image ($T=1$) or video ($T>1$) sequence by downsampling the input sequence $\mathcal{F}$ according to the predefined target resolutions $\{(H_\ell, W_\ell)\}_{\ell=0}^{L}$ (Line 1). 
If the input is a video sequence ($T>1$), I-frames are selected using Dynamic Key-frame Selection (Lines 2--4).

For each frame $f_t$ (Line 5--11), the initialization procedure differs depending on whether the input is a video and the current frame is an I-frame or the input is an image.
If $f_t$ is an I-frame or if the input is an image, we initialize the layered Gaussian splats 
$\mathcal{G}_t = \{\mathcal{G}_t^{0}, \Delta \mathcal{G}_t^{1}, \ldots, \Delta \mathcal{G}_t^{L}\}$ 
with $\boldsymbol{N} +\boldsymbol{N}_{\text{aug}}$ random Gaussian splats (Line 6--7). 
Otherwise, we initialize $\mathcal{G}_t$ using the splats from the previous frame $\mathcal{G}_{t-1}$ and inject $\boldsymbol{N}_{\text{aug}}$ random Gaussian splats (Line 8--10), distributed evenly across all layers.

At each iteration $k$ (Line 12), a target resolution level $\ell_k$ is cyclically selected according to $\ell_k \equiv k  \big(\mathrm{mod} (L-1)\big)$ (Line 13). 
The rendered output at resolution level $\ell_k$ is obtained via rendering of the first $\ell_k$ Gaussian layers according to~\Cref{equ:render} (Line 14). 
A joint loss is computed based on~\Cref{equ:k-L}, and the layered Gaussian parameters are updated accordingly (Line 15).
To reduce redundancy, Gaussian splats with minimal contribution (measured by importance weight $w$ in~\Cref{equ:rendering_rgbW}) are periodically removed. 
Specifically, for every $Z_{\mathrm{GSP}}$ iterations during the first $K_{\mathrm{GSP}}$ iterations, 
$\boldsymbol{N}_{\mathrm{prune}} \cdot \frac{Z_{\mathrm{GSP}}}{K_{\mathrm{GSP}}}$ splats are pruned (Line 16). 
The optimization continues until convergence for the current $t$-th frame (Line 12--17).
Finally, the optimized layered Gaussian representation is obtained as 
$\mathcal{G} = \{\mathcal{G}^{0}, \Delta \mathcal{G}^{1}, \ldots, \Delta \mathcal{G}^{L}\}$.

\subsection{Quantization~\label{sec:quantization}}
We adopt the temporal-aware fine-tuning and quantization strategy of GSVC~\cite{DBLP:conf/nossdav/WangSO25}, while maintaining the layered structure of progressive splatting and employing joint optimization.
For I-frames, all Gaussian parameters in both the base and enhancement layers are directly quantized.
For P-frames, we quantize the parameter differences relative to the reference frame.

To quantize each parameter:
The position $\boldsymbol{\mu}$ is stored in reduced floating-point precision~\cite{DBLP:conf/aaai/ZhuLCZJWY25}.
Specifically, we use 16-bit float for I-frames and 12-bit for P-frames.
For the Cholesky vector $\boldsymbol{\ell}$, we apply $b$-bit asymmetric quantization~\cite{DBLP:journals/corr/abs-2004-09576}: 
\begin{equation}
\begin{array}{c} 
    \mathcal{Q}_{\ell}(\ell_i) = \hat{\ell}_i \,\gamma_i + \beta_i \\
    \hat{\ell}_i = \left\lfloor 
        \text{clamp}\!\left(
            \dfrac{\ell_i - \beta_i}{\gamma_i},\ 
            L_{\min},\ L_{\max}
        \right) 
    \right\rfloor,\\
    L_{\max} - L_{\min} = 2^{b} - 1.
\end{array}
\end{equation}
where $i \in \{0,1,2\}$, and $\gamma_i$ and $\beta_i$ denote the learnable scaling and offset stored in 32-bit floating-point format.
For I-frames, the base layer is quantized as unsigned integers with 
$[L_{\min}, L_{\max}] = [0, 2^{b}-1]$, 
while the enhancement layers are quantized as signed integers with 
$[L_{\min}, L_{\max}] = [-2^{b-1}, 2^{b-1}-1]$, 
where $b$ is set to 6. 
For P-frames, both the base and enhancement layers are quantized as signed integers with 
$[L_{\min}, L_{\max}] = [-2^{b-1}, 2^{b-1}-1]$, 
where $b$ is set to 5.
For the weighted color $\boldsymbol{c}$, residual vector quantization (VQ)~\cite{DBLP:journals/taslp/ZeghidourLOST22} is employed, stacking $M$ stages of VQ~\cite{gray1984vector} with codebook size $B$:
\begin{equation}
\begin{array}{c} 
   \mathcal{Q}_{c}(\boldsymbol{c}') = \sum_{k=1}^M \boldsymbol{C}^k[i^k],\\ 
   i^k = \arg\min_{k} \left\| \boldsymbol{C}^k[k] - \left(\boldsymbol{c}' - \sum_{j=1}^{k-1} \boldsymbol{C}^j[i^j]\right)\right\|_2^2,
\end{array}
\end{equation}
where $\boldsymbol{C}^k \in \mathbb{R}^{B \times 3}$ is the $k$-th codebook and $i^k$ its index.
For both I-frame and P-frames, $B$ is set to 6-bit.
The codebooks are optimized with a commitment loss~\cite{DBLP:conf/eccv/ZhangGXHWQLGZ24}.

\section{Experiments}
\subsection{Experimental Setup}

\textbf{Dataset.}
For image experiments, we adopt the Kodak dataset~\cite{KodakDataset}, which contains 24 natural images with a resolution of $768 \times 512$, and the DIV2K validation set~\cite{DBLP:conf/cvpr/AgustssonT17}, which consists of 100 high-quality images at the 2K level.
These two datasets cover diverse levels of resolution for the image reconstruction scenarios.
For video experiments, we use all 1080p videos from the UVG dataset~\cite{DBLP:conf/mmsys/MercatVV20}.
Each video is standardized to $1920 \times 1080$ resolution, 4:2:0 chroma subsampling, 8-bit depth, and YUV color space.
In our experiments, we use the first 50 frames of each video.  

We run two sets of experiments, one for resolution scalability and the other for quality scalability.
For resolution scalability,  we generate images and videos with multiple resolution levels by downsampling the images and videos in the datasets based on the number of splats $N^i$, with the ratio for level~$l$ defined as $(\sum_{i=0}^l N^i/N)^{1/2}$.  
The ground truth for evaluation at each level matches the reduced resolution of the rendered output.
For quality scalability, the ground truth used for evaluation at each level is the full-resolution version.

\textbf{Implementation.}
Our {\name} framework is implemented on the gsplat library~\cite{DBLP:journals/corr/abs-2409-06765} with CUDA-based rasterization and accumulated blending~\cite{DBLP:conf/eccv/ZhangGXHWQLGZ24}. Gaussian parameters are optimized using the Adan optimizer~\cite{DBLP:journals/pami/XieZLLY24} with an initial learning rate of $1\times10^{-3}$, reduced by half every 20,000 steps.
We adopt three layers ($L=3$) with splats evenly distributed across them ($\boldsymbol{N}=[\frac{N}{3},\frac{N}{3},\frac{N}{3}]$), and apply constant-rate coding by varying the total number of splats $N$ from 12k to 27k with a step of 3k.
The removal ratio is set to 0.2 for the base layer and 0.4 for enhancement layers ($\boldsymbol{N}_{aug} = \boldsymbol{N}_{prune} 
= \boldsymbol{N} \odot [0.2, 0.4, 0.4]$).
For I-frame, $Z_{\mathrm{GSP}}$ is set to 4,000, for P-frames, $K_{\mathrm{GSP}}$ is set to 1,000.
$Z_{\mathrm{GSP}}$ is set to 100.
Other settings of DKS, GSP, and GSA follow GSVC~\cite{DBLP:conf/nossdav/WangSO25}.
Training continues until convergence, defined as 100 iterations with less than $1\times10^{-7}$ change in joint loss, and all experiments are conducted on an NVIDIA H100 GPU.

\begin{figure*}[t]
    \centering
    \subfloat[Quality Scalablility]{\includegraphics[width=0.48\textwidth]{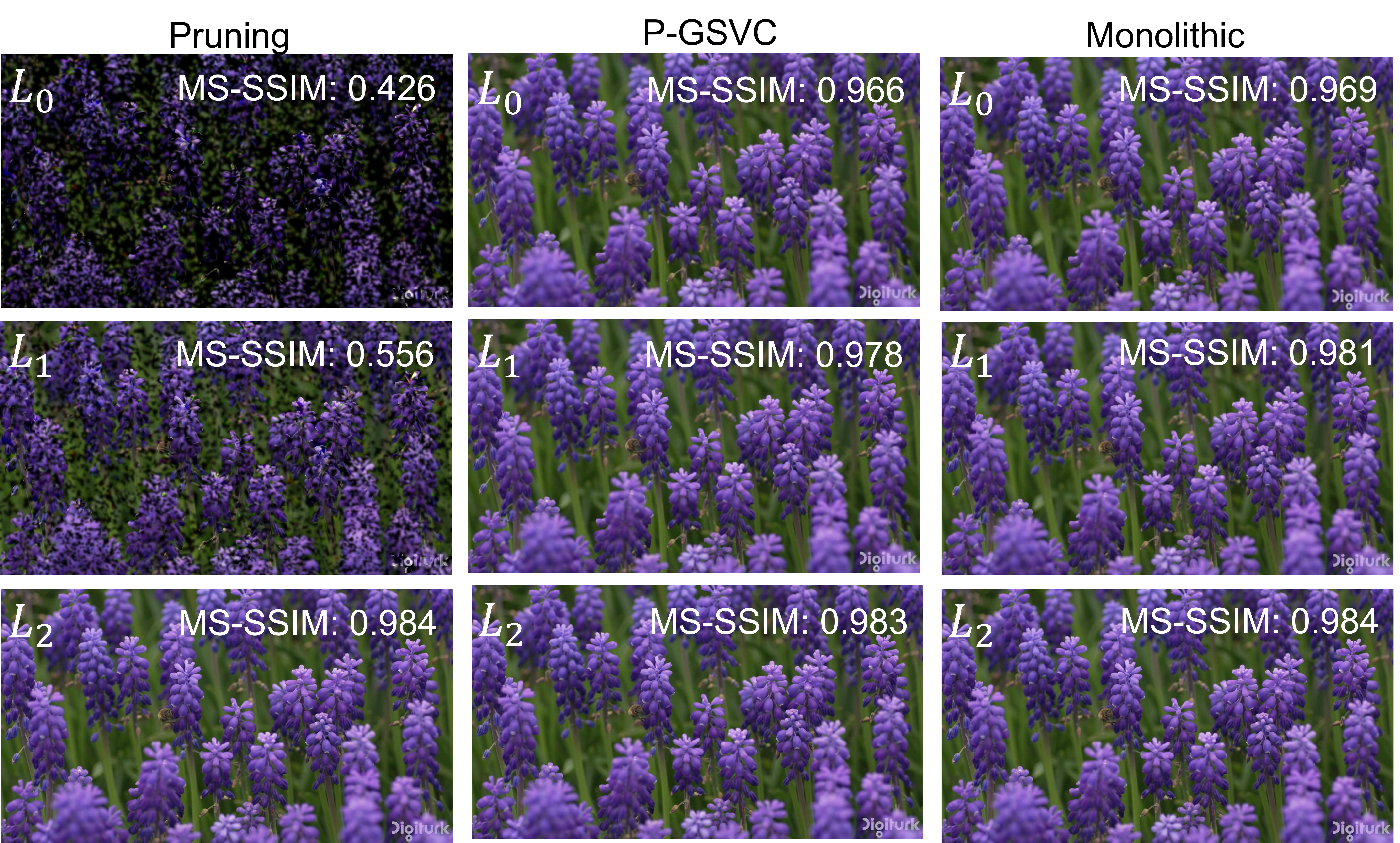}}\hspace{1px}
    \subfloat[Resolution Scalability]{\includegraphics[width=0.48\textwidth]{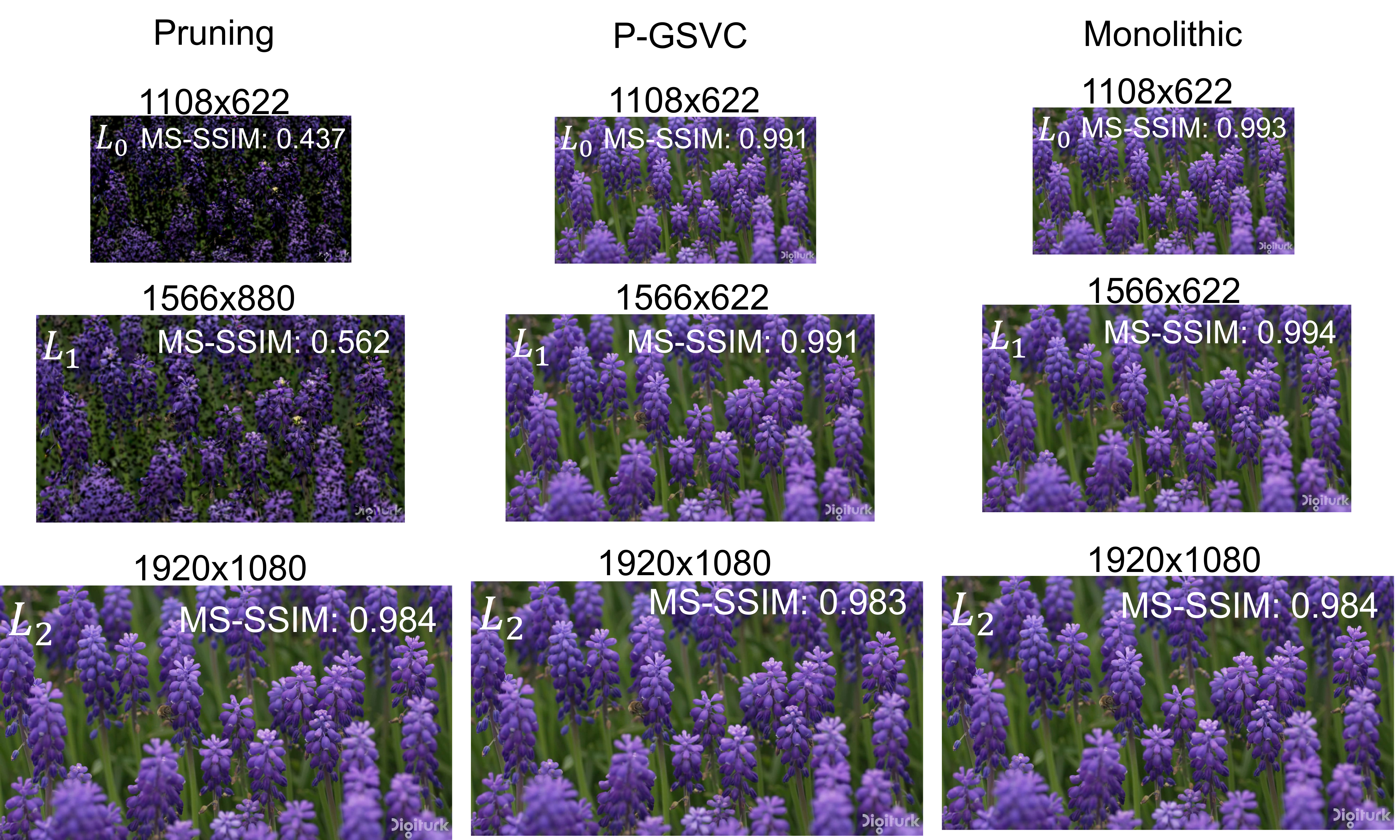}}\hspace{1px}
    \caption{
    A sampled frame rendering from \textit{HoneyBee} at three levels, $L_0$, $L_1$, and $L_2$. (a) shows quality scalability, and (b) shows resolution scalability.~\name~eliminates the artifacts of the \textsf{Pruning} method, achieving results comparable to the \textsf{Monolithic} method.}
    \label{fig:scalability_rendering}
\end{figure*}

\textbf{Comparison Method.} 
We conducted separate comparisons between our~\name~and three alternative methods:
\begin{itemize}[leftmargin=*]
    \item \textsf{Pruning.} A single full-scale Gaussian model is trained, and lower levels are obtained by pruning splats according to their contribution based on $\boldsymbol{w}$ in~\Cref{equ:rendering_rgbW}. This naive method serves as a naive baseline for scalable coding.
    \item \textsf{Monolithic.} Separate Gaussian models are trained independently for each level with increasing Gaussian counts. 
    This approach is not scalable, as each level requires a separate set of splats rather than being constructed progressively. 
    In our experiments, this serves as an upper bound in quality.
    
    \item \textsf{Sequential.} A layered Gaussian model is trained sequentially: parameters from the lower layers are frozen and passed to the next higher layer, where enhancement splats are added and optimized. This strategy adopts the same layered structure as \name~but trains layers independently without joint optimization. It is commonly used in layered 3DGS~\cite{DBLP:conf/3dim/ShiMGO25} and 2DGS~\cite{DBLP:conf/aaai/ZhuLCZJWY25}, and serves as a state-of-the-art baseline for comparison with our joint training strategy.
\end{itemize}

To comprehensively demonstrate the effectiveness of~\name, we conduct three groups of experiments in image and video cases.
First, we evaluate our joint training strategy on image datasets, where the absence of temporal dependency enables an isolated comparison with sequential layer-wise training.
In particular, for the \textsf{Sequential} method, we adopt LIG~\cite{DBLP:conf/aaai/ZhuLCZJWY25}, the state-of-the-art framework for layered 2D Gaussian training on images, which also follows a sequential optimization scheme. For fairness, all methods use the same three-level configuration with equal Gaussian budgets per layer, while keeping other parameters identical.
Second, we evaluate scalability on video datasets. Since an image can be regarded as a special case of video (a single I-frame sequence), we extend our framework to full video sequences to examine both quality and resolution scalability. We compare \textsf{Pruning}, \textsf{Monolithic}, \textsf{Sequential}, and our {\name} under identical settings.
Finally, we evaluate compression performance under quantization on video datasets by examining the rate-distortion trade-off, with the \textsf{Monolithic} model serving as an upper bound and {\name} compared against the \textsf{Sequential} baseline.

\textbf{Metrics.} 
We evaluate reconstruction quality using Peak Signal-to-Noise Ratio (PSNR) and Multi-Scale Structural Similarity (MS-SSIM).
PSNR, expressed in decibels (dB), quantifies pixel-level fidelity by comparing signal power to reconstruction error, where higher values indicate better quality.
MS-SSIM measures perceptual similarity across multiple scales with values ranging from 0 to 1, where values closer to 1 indicate better structural similarity.
For the image case, we additionally employ the Learned Perceptual Image Patch Similarity (LPIPS) metric~\cite{DBLP:conf/cvpr/ZhangIESW18}, which evaluates perceptual similarity by comparing deep feature representations extracted from pretrained neural networks. LPIPS values typically range from 0 to 1, where lower values indicate higher perceptual similarity and better visual quality.
For video, we additionally employ the Video Multi-Method Assessment Fusion (VMAF) metric~\cite{li2018vmaf}, 
a learning-based perceptual quality model that combines multiple quality indices to predict human subjective assessment. 
VMAF scores typically range from 0 to 100, with higher values indicating better perceived visual quality.

\begin{table}
\centering
\caption{Comparison of visual quality of {\name} and \textsf{Sequential} method (LIG) on the Kodak and DIV-HR datasets with different numbers of splats.}
\label{tab:P-GSVC_vs_LIG}
    \resizebox{\linewidth}{!}{ 
    \begin{tabular}{cc|ccc|ccc}
        \toprule
        \multicolumn{2}{c|}{\text{Dataset}} & \multicolumn{3}{c|}{\text{{Kodak}}} &\multicolumn{3}{c}{\text{{DIV-HR}}}\\
        \multicolumn{2}{c|}{\text{No. Splats}} &$5\times10^3$&$7\times10^3$&$9\times10^3$&$5\times10^4$&$7\times10^4$&$9\times10^4$\\
        \midrule
        \multirow{2}{*}{\text{{PSNR $\uparrow$}}}&\text{LIG}&26.6 &27.3 &28.1 &27.8 &28.7 &29.5 \\
        &\text{\name}&\textbf{28.5} &\textbf{29.4} &\textbf{30.2} &\textbf{30.2} &\textbf{31.3} &\textbf{32.1} \\
        \midrule
        \multirow{2}{*}{\text{{MS-SSIM $\uparrow$}}}&\text{LIG}&0.918 &0.937 &0.948 &0.944 &0.958 &0.966 \\
        &\text{\name}&\textbf{0.943} &\textbf{0.957} &\textbf{0.965} &\textbf{0.967} &\textbf{0.975} &\textbf{0.981} \\
        \midrule
        \multirow{2}{*}{\text{{LPIPS $\downarrow$}}}&\text{LIG}&0.436 &0.380 &0.339 &0.383 &0.325 &0.284 \\
        &\text{\name}&\textbf{0.271} &\textbf{0.223} &\textbf{0.189} &\textbf{0.217} &\textbf{0.178} &\textbf{0.150} \\
        \bottomrule
    \end{tabular}
    }
\end{table}

\subsection{Results and Analysis}

\textbf{Image.} 
We evaluate the joint training strategy of~\name~on image representation tasks against the sequentially layer-wise training method on the Kodak and DIV-HR datasets. 
Following the setup described earlier, we compare~\name~against the \textsf{Sequential} method implemented by LIG~\cite{DBLP:conf/aaai/ZhuLCZJWY25}—the current state-of-the-art framework for layered 2D Gaussian training on images.
LIG~\cite{DBLP:conf/aaai/ZhuLCZJWY25} adopts a layered Gaussian structure by sequentially training a skeleton layer followed by residual layers.    The results is summarized in Table~\ref{tab:P-GSVC_vs_LIG}.

We first report on the highest-level image reconstruction performance on the Kodak and DIV-HR datasets, under varying total Gaussian budgets.
The performance of individual layers and their progressive contributions will be further examined in the following experiments.

The results show that, 
on both the Kodak and DIV-HR datasets,~\name~achieves higher PSNR and MS-SSIM across all Gaussian budgets, confirming the advantage of joint training in optimizing layered Gaussian representations.
Specifically, {\name} improves PSNR by approximately 1.9--2.1 dB on Kodak and 2.4--2.6 dB on DIV-HR, with consistent gains in MS-SSIM and LPIPS.
Overall, ~\name~consistently outperforms LIG, demonstrating the effectiveness of our joint training strategy over the sequential layer-wise approach in image representation tasks.

\textbf{Video.}  
Since an image can be regarded as a single I-frame sequence, we further extend our experiments to video datasets to evaluate the quality and resolution scalability of~\name.

We first present qualitative comparisons for the three levels' output. 
As shown in~\Cref{fig:scalability_rendering}, in both quality and resolution scalability tasks, when only partial splats are decoded in $L_0$ and $L_1$, the \textsf{Pruning} method produces severe artifacts such as holes and broken structures.
These missing regions directly reflect that the splats responsible for coarse structures are discarded when ranking solely by contribution.
The contribution score only reflects the overall importance of each Gaussian but cannot disentangle whether it carries coarse structures or fine details. As a result, selecting only the top-contributing splats fails to preserve complete reconstructions at lower levels, leading to severely degraded visual quality.
In contrast, \name~organizes splats into layered structures and jointly optimizes them, leading to artifact-free intermediate reconstructions in $L_0$ and $L_1$.
Moreover, when compared with the \textsf{Monolithic} method, \name~achieves comparable visual quality across all layers, demonstrating that it preserves scalability with a minor quality sacrifice.

As \textsf{Pruning} produces severely degraded intermediate quality with artifacts, and thus cannot serve as a meaningful baseline, it is excluded from further experiment results below.

\begin{figure*}[t]
\centering
    \begin{minipage}[b]{0.95\textwidth}\centering
      \subfloat[Quality Scalablility on UVG]{
        \includegraphics[width=0.29\textwidth]{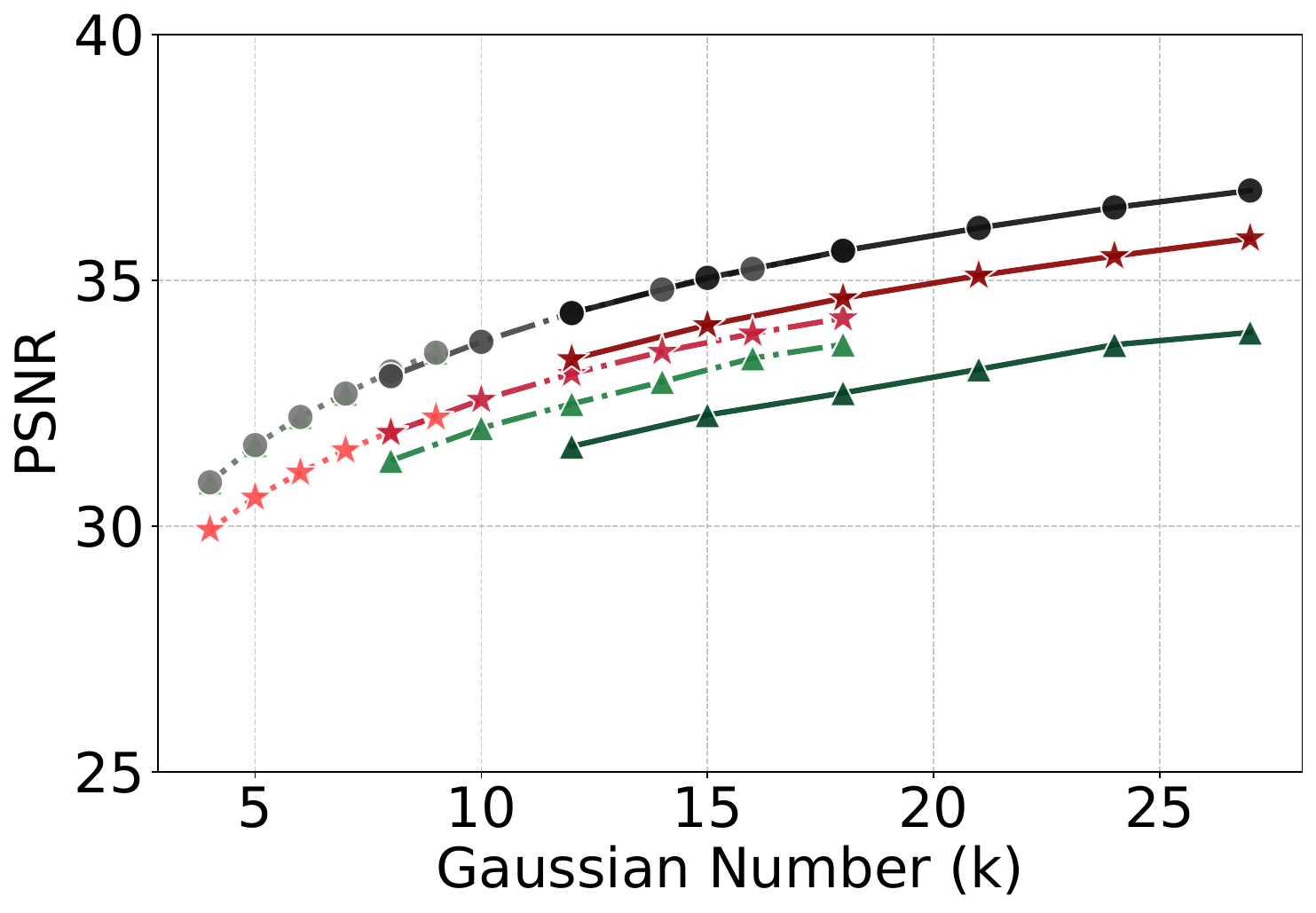}
        \includegraphics[width=0.29\textwidth]{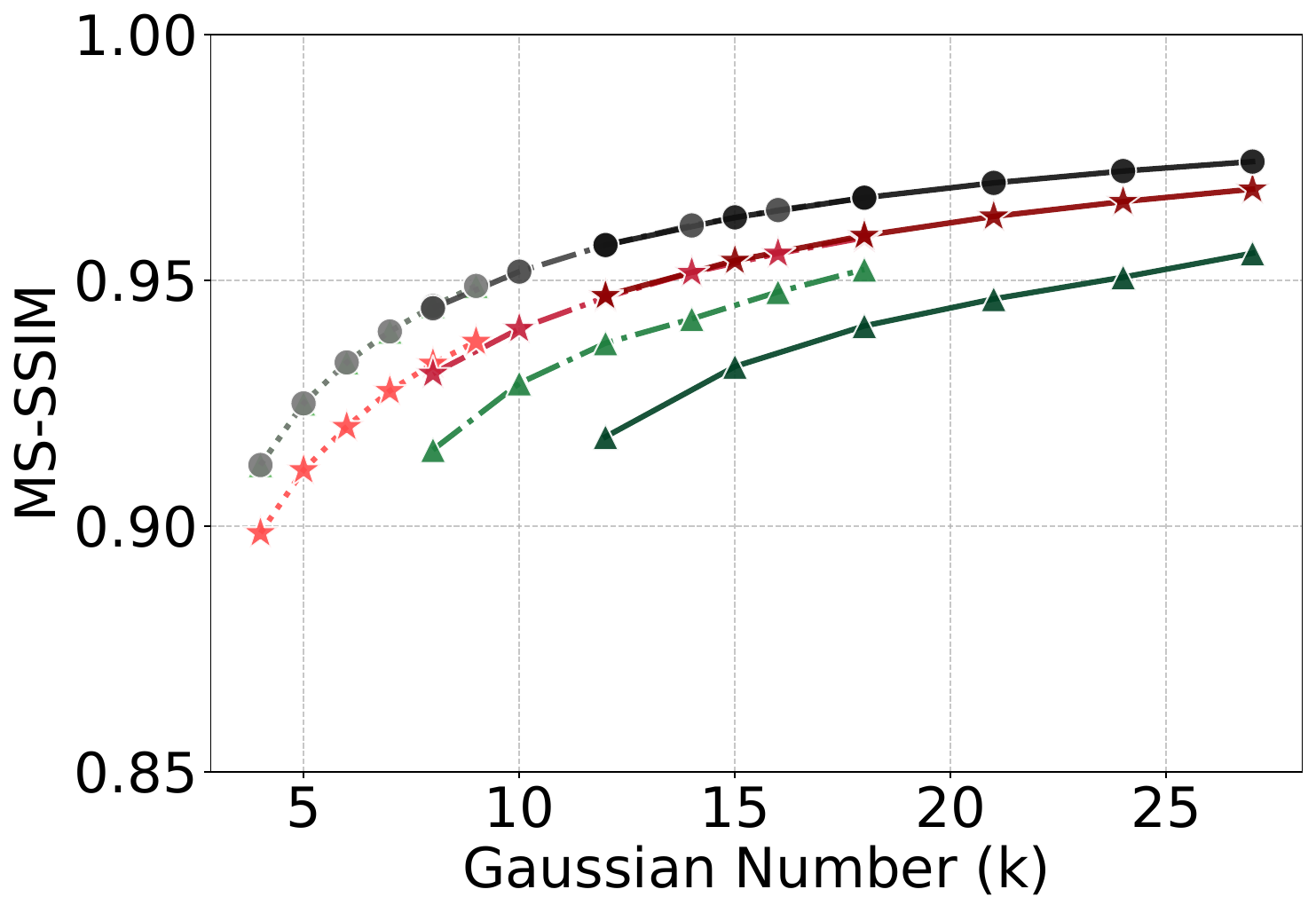}
        \includegraphics[width=0.29\textwidth]{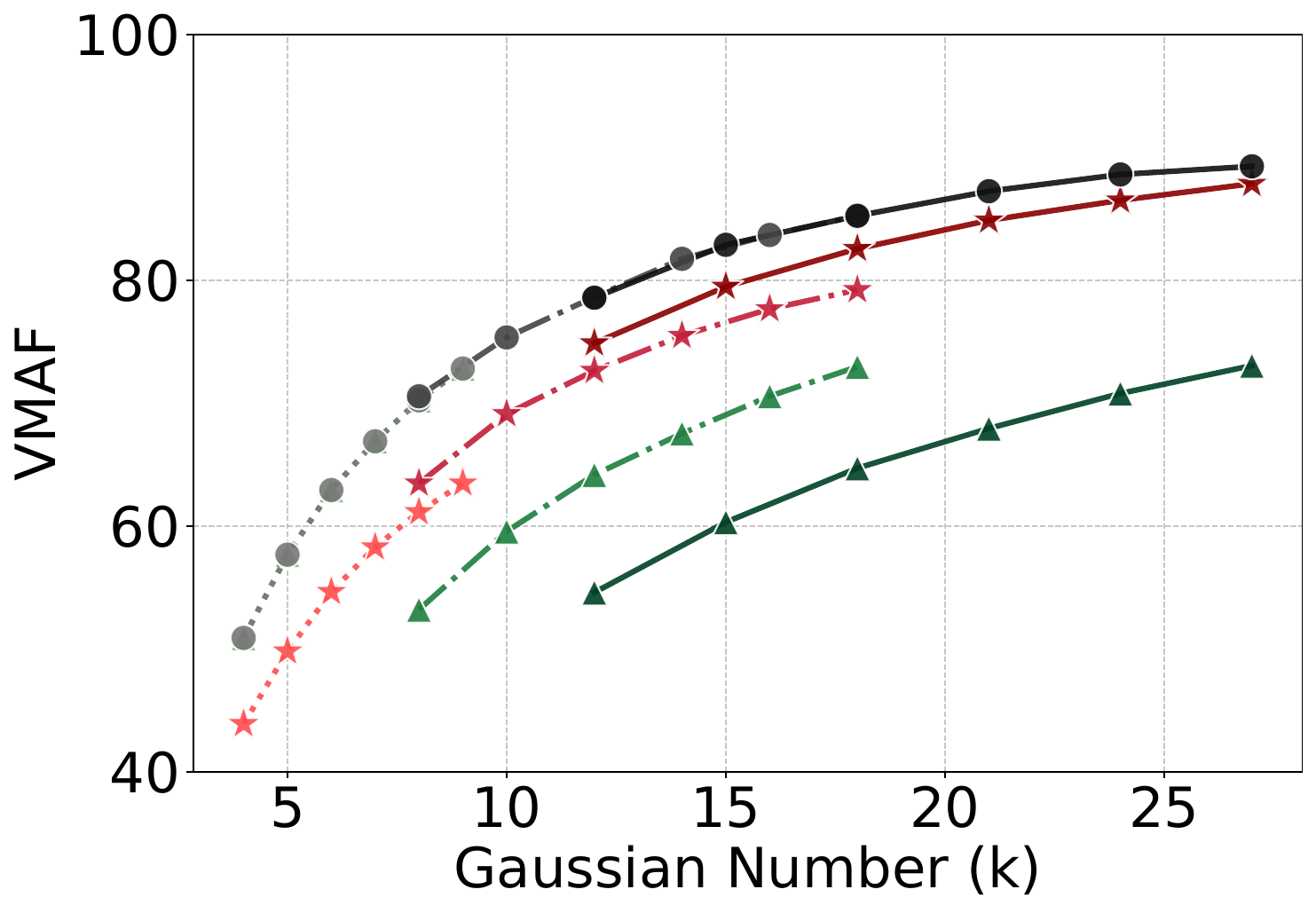}
        \raisebox{5mm}{\includegraphics[width=0.13\textwidth]{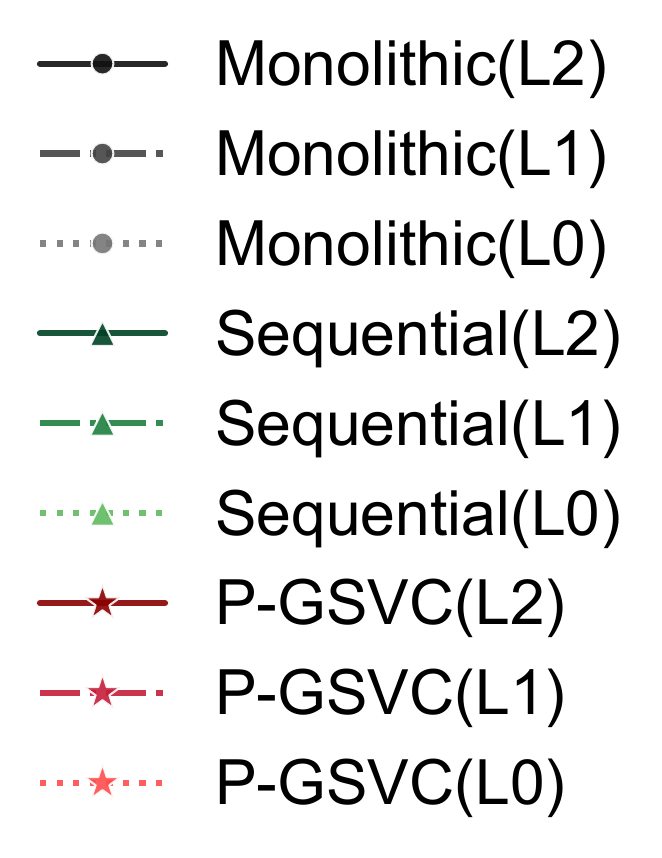}}
        \label{fig:quality_scalability_on_UVG}
      }\hfill
    \vspace{0.5em}
    \end{minipage}
    \begin{minipage}[b]{0.95\textwidth}\centering
      \subfloat[Resolution Scalability on UVG]{
        \includegraphics[width=0.29\textwidth]{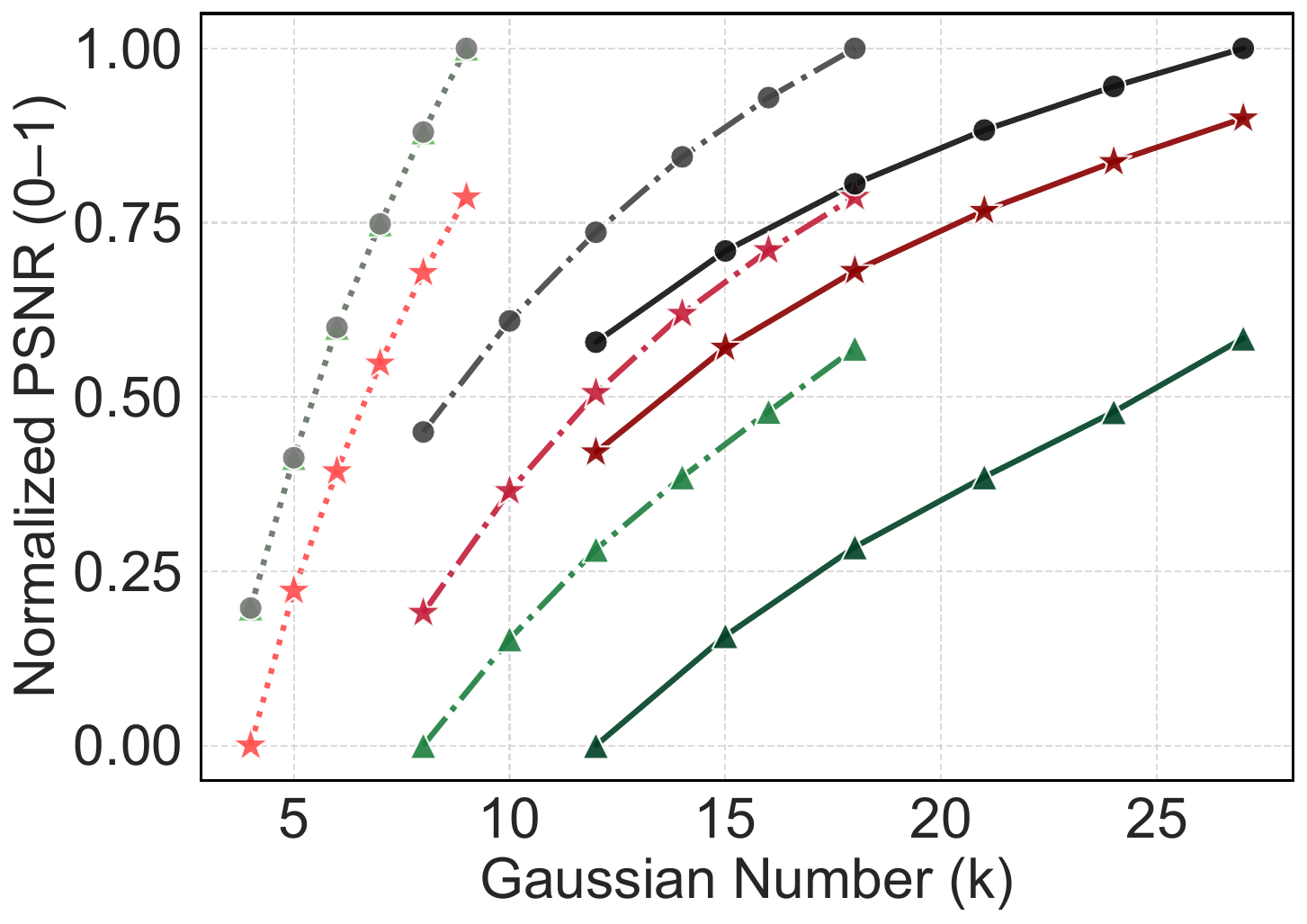}
        \includegraphics[width=0.29\textwidth]{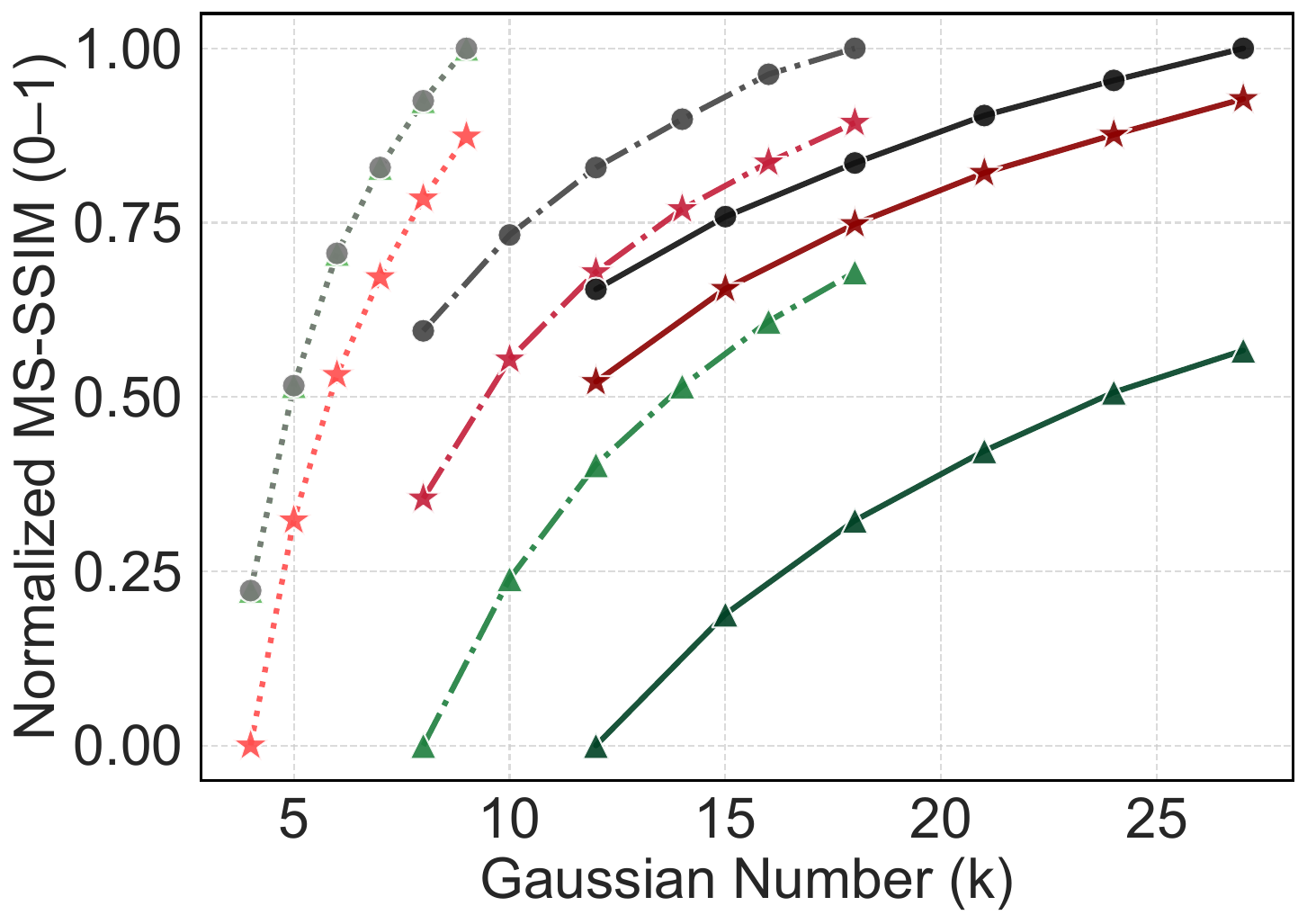}
        \includegraphics[width=0.29\textwidth]{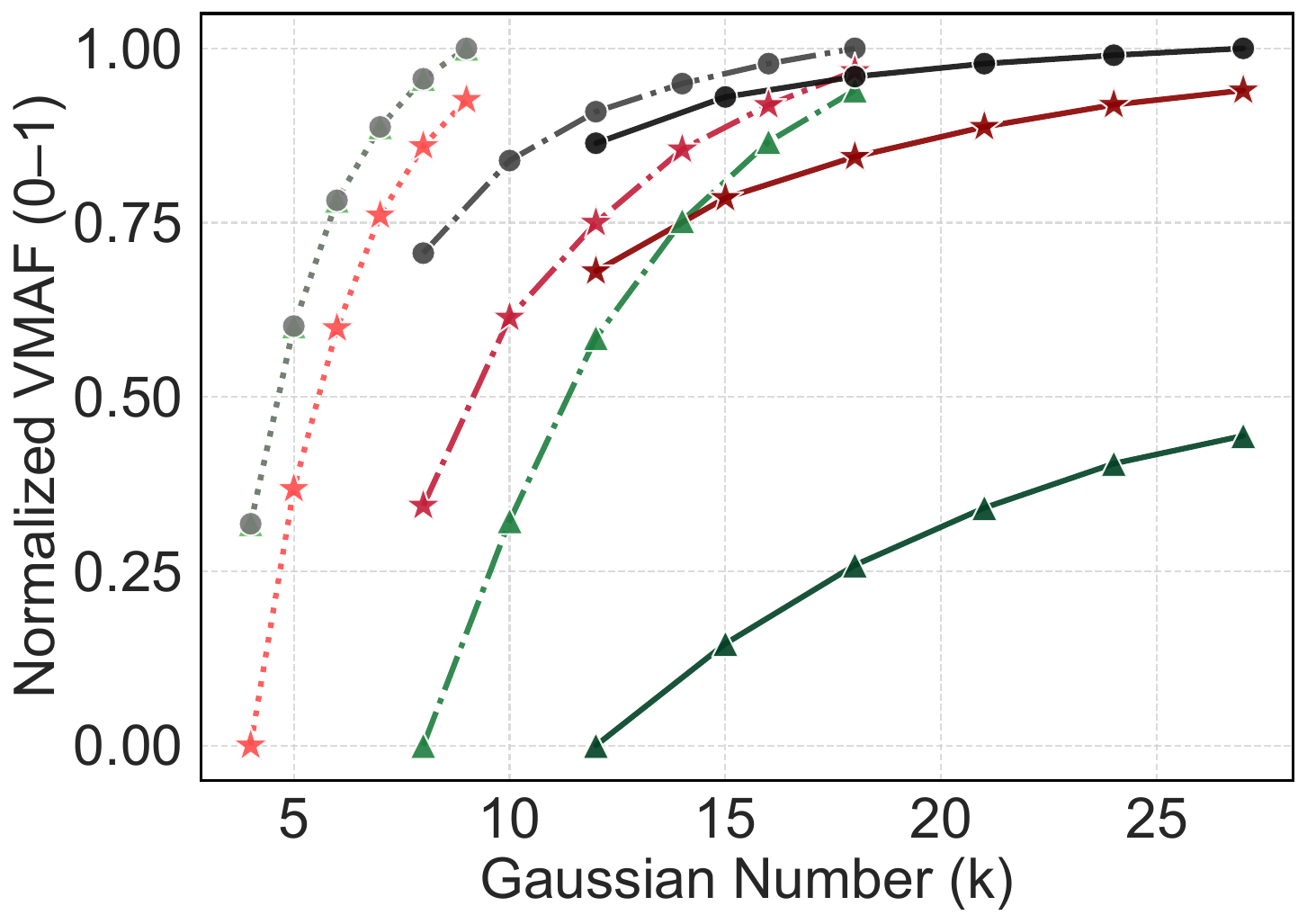}
        \raisebox{5mm}{\includegraphics[width=0.13\textwidth]{Fig/Experiment/legend_UVG.pdf}}
        \label{fig:resolusion_scalability_on_UVG}
      }\hfill
    \end{minipage}
\caption{Scalability evaluation on the UVG dataset, comparing (a) quality and (b) resolution across \textsf{Monolithic}, \textsf{Sequential}, and {\name} methods.}
\label{fig:scalability_on_UVG}
\end{figure*}

Next, we present the experimental results on the scalability of \name~in terms of PSNR, MS-SSIM, and VMAF on the UVG dataset, considering both quality scalability and resolution scalability illustrated in~\Cref{fig:scalability_on_UVG}.

\Cref{fig:quality_scalability_on_UVG} reports the quality scalability results on the UVG dataset in terms of PSNR and MS-SSIM. The x-axis denotes the number of splats, which is positively correlated with model size, while the y-axis indicates reconstruction quality. The results reveal three key observations:  
(i) For all methods, higher layers consistently improve quality as more splats are added.  
(ii) For the \textsf{Sequential} approach, the base layer achieves a similar quality to the \textsf{Monolithic} method because its optimization is unconstrained by higher layers and thus follows the same training process as the upper bound. However, despite this relatively strong base layer, the enhancement layers cannot be effectively optimized due to conflicting optimization objectives across layers.  As a result, the training falls into a suboptimal local minima, which is reflected in the small quality gains across layers. For example, as shown, the top quality level of the \textsf{Sequential} method remains far below the $L_1$ of the \textsf{Monolithic} method, and the gap between $L_2$ exceeds 2.2 dB. 
(iii) In contrast, \name~achieves more efficient scalability than the \textsf{Sequential} method. Although its base layer is weaker (< 3\% lower), it shows significantly larger improvements at higher levels, with $L_1$ exceeding the \textsf{Sequential} method by about 0.5 dB and $L_2$ by nearly 2 dB. At the same time, the gap to \textsf{Monolithic} is substantially reduced, demonstrating that \name~preserves scalability while incurring a smaller quality sacrifice.

\Cref{fig:resolusion_scalability_on_UVG} shows the resolution scalability results on the UVG dataset.
In resolution scalability, each level is evaluated against ground truth at its corresponding resolution, with higher levels having higher resolutions.
As a result, absolute PSNR, MS-SSIM, or VMAF values are not directly comparable across levels; therefore, results are normalized within each level to [0, 1].
For a metric value $m_{l}$ at level $l$, we apply min-max normalization:
\begin{equation}
\tilde{m}_{l} = 
\frac{m_{l}-m^{\min}_{l}}{m^{\max}_{l}-m^{\min}_{l}},
\quad
\end{equation}
where $m^{\min}_{l}$ and $m^{\max}_{l}$ denote the minimum and maximum metric values, respectively, computed over all evaluated methods and Gaussian configurations within level $l$.
In this figure, the x-axis denotes the number of splats, while the y-axis represents the normalized quality score.
Compared with the \textsf{Sequential} method, our method yields substantially larger gains in the enhancement layers, achieving over 0.25 normalized VMAF gain at $L_1$ and nearly 0.5 at $L_2$, while at the same time narrowing the gap to the \textsf{Monolithic}.
These results show that our~\name~achieves more effective resolution scalability than the \textsf{Sequential} approach.

\begin{figure*}[t]
    \centering
    \includegraphics[width=0.29\textwidth]{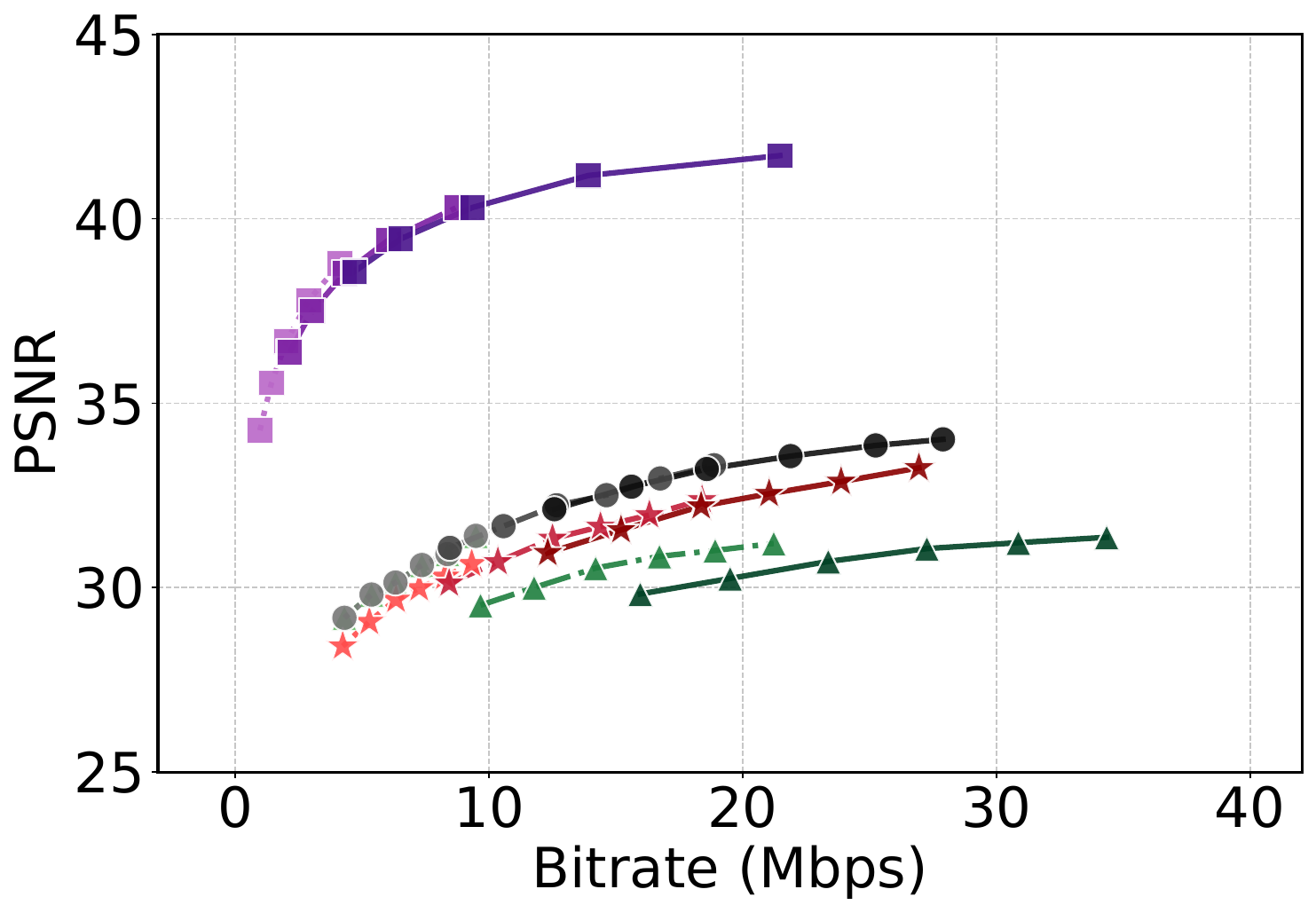}
    \includegraphics[width=0.29\textwidth]{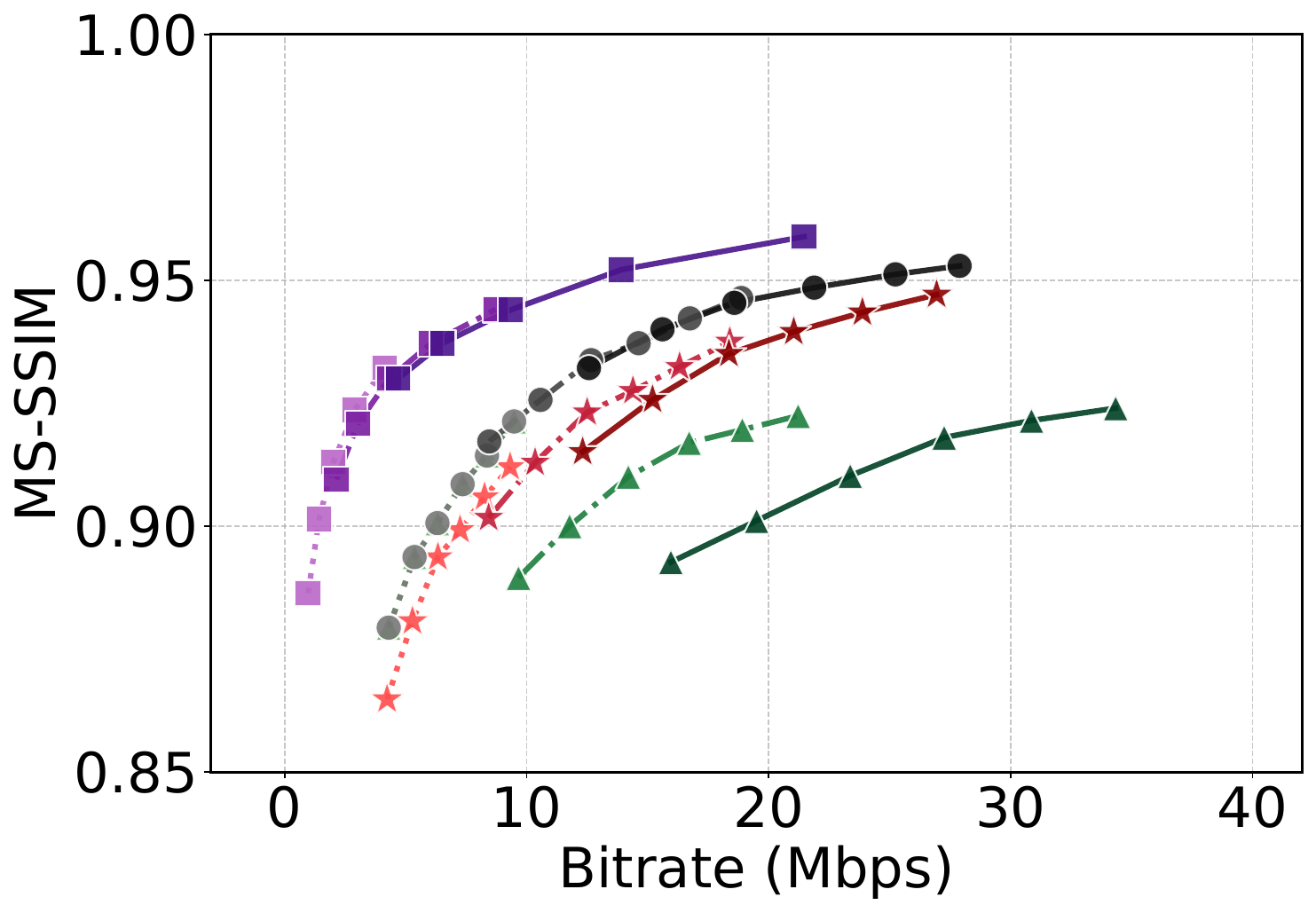}
    \includegraphics[width=0.29\textwidth]{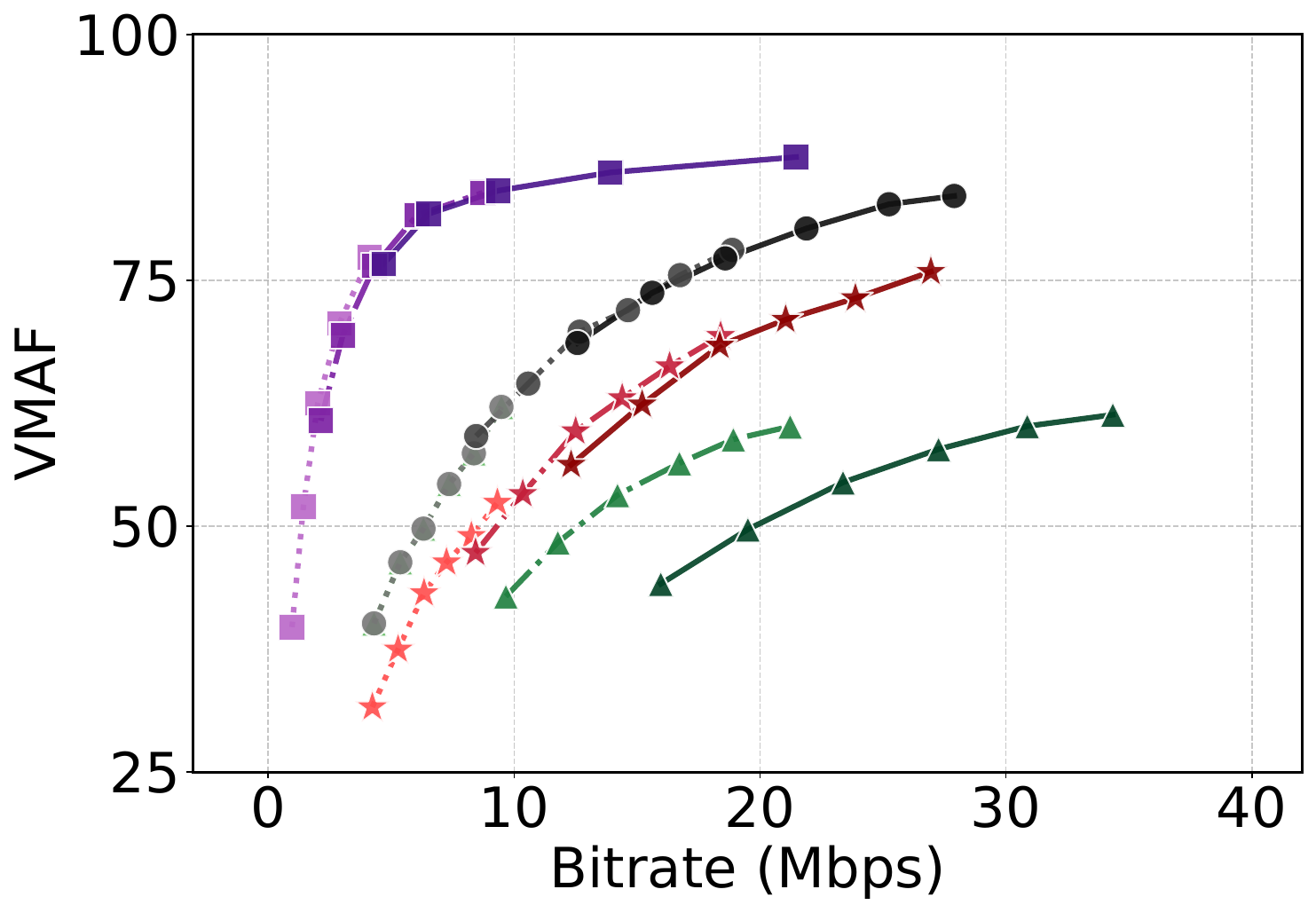}
    \raisebox{5.5mm}{\includegraphics[width=0.11\textwidth]{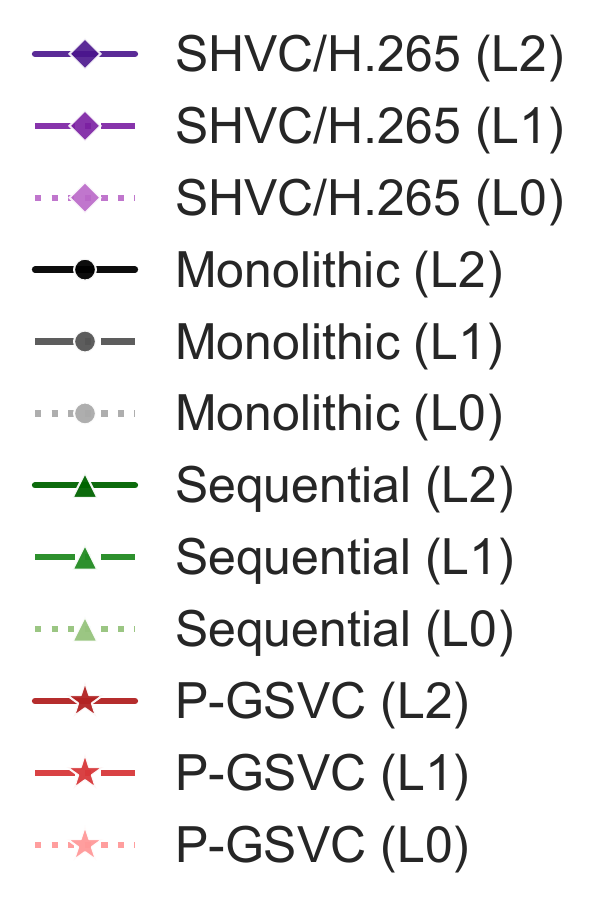}}
    \caption{Quantized rate–distortion results of~\name~versus baselines on UVG dataset.}
    \label{fig:quantization}
\end{figure*}

\textbf{Rate-Distortion Trade-off.}  
\Cref{fig:quantization} presents the rate--distortion results on the UVG dataset, comparing our~\name~with the \textsf{Monolithic} and \textsf{Sequential} methods after quantization, while also including SHVC~\cite{DBLP:journals/tcsv/BoyceYCR16} in a three-level setting.
In this comparison, SHVC is run with the well-optimized SHM-12.4 reference software, which gives it stronger rate–distortion performance and makes it a solid baseline for standardized scalable video coding.
The \textsf{Monolithic} method trains each level separately without parameter sharing.
Since it does not perform layer-wise training, it does not have conflicting cross-layer optimization objectives or stability issues when switching layers.  
In the quality scalability setting, it is essentially equivalent to GSVC evaluated under different Gaussian budgets.
This strategy is not scalable, as each level is trained independently rather than derived from a single bitstream; in our experiments, it serves as the upper bound to quantify the quality overhead of achieving scalability in~\name.
On the contrary, the \textsf{Sequential} method achieves scalability with parameter sharing across levels.
However, its layer-wise training 
results in considerable overhead and performance degradation, due to conflicting optimization objectives as a new layer is added.

Our~\name~employs a layered structure in which Gaussian parameters are shared across levels to achieve scalable video.
As a result, the rate-distortion curves appear piecewise, yet the gaps between adjacent levels remain very small (around 0.2 dB), demonstrating the effectiveness of joint training. 
Inevitably, achieving scalability introduces a quality overhead.
Under our default strategy of evenly distributing splats across layers, the quality remains below about 1.1~dB in PSNR across all layers under the same Gaussian budget.  The difference in quality reduces at higher quality levels as the total number of splats increases.
This trend suggests that allocating more splats to enhancement layers, which focus on reconstructing high-frequency details, can further reduce the overhead while preserving scalability.  

While our current implementation does not match the performance of the highly optimized modern standard (e.g., SHVC), we narrow the gap between SHVC and the \textsf{Sequential} method, demonstrating meaningful progress toward standardized scalable coding quality.  Additionally, note that P-GSVC uses an L2 loss for optimization, and by tuning the convergence condition, we can further improve PSNR without increasing the number of splats.

\textbf{Size and Computational Cost}.
For a 600-frame YUV video (1920$\times$1080, about 1.9\,GB in raw format), encoding all three layers with {\name} produces an average total bitstream of about 45--100\,MB per video under the per-frame Gaussian budgets of 12k--27k. 
The encoding time of our implementation of {\name} takes around 720 seconds per frame. 
Meanwhile, {\name} achieves real-time rendering, simultaneously reconstructing all three layers at a speed of about 1200 fps (2 s.f.). 
Although the encoding cost is prohibitively high, in offline settings such as video-on-demand, encoding is done once, and the cost is amortized over multiple viewings. 
Moreover, our implementation has not been optimized for speed.  Profiling shows that over 90\% of computation lies in iterative Gaussian optimization, which can be accelerated through parallelization on multiple GPUs (a potential 5.33$\times$ speedup on multi-GPU reported in~\cite{zhao2024scaling}).  We are currently developing a faster version of the encoder for practical deployment.
\section{Conclusion}
In this paper, we introduced {\name}, the first scalable Gaussian representation framework for both images and videos, built upon progressive 2D Gaussian Splatting.  We highlighted and addressed a challenge in optimizing progressive Gaussian representations, showing that incremental layer-wise training would lead to poorer quality due to cross-layer optimization conflicts.  Through joint, incremental, and cyclic training across levels, {\name} demonstrates that we can achieve stable progressive reconstruction of both images and videos with significant improvements over layer-wise sequential training.
This work positions Gaussian splats as an alternative primitive for encoding images and videos for adaptive media delivery, bridging between classical scalable codecs and neural-based codecs.


\bibliographystyle{ACM-Reference-Format}
\balance
\bibliography{ref}

@article{DBLP:journals/tcsv/BoyceYCR16,
  author       = {Jill M. Boyce and
                  Yan Ye and
                  Jianle Chen and
                  Adarsh K. Ramasubramonian},
  title        = {Overview of {SHVC:} Scalable Extensions of the High Efficiency Video
                  Coding Standard},
  journal      = {{IEEE} Trans. Circuits Syst. Video Technol.},
  volume       = {26},
  number       = {1},
  pages        = {20--34},
  year         = {2016},
  url          = {https://doi.org/10.1109/TCSVT.2015.2461951},
  doi          = {10.1109/TCSVT.2015.2461951},
  bibsource    = {dblp computer science bibliography, https://dblp.org}
}

@article{DBLP:journals/tcsv/SchwarzMW07,
  author       = {Heiko Schwarz and
                  Detlev Marpe and
                  Thomas Wiegand},
  title        = {Overview of the Scalable Video Coding Extension of the {H.264/AVC}
                  Standard},
  journal      = {{IEEE} Trans. Circuits Syst. Video Technol.},
  volume       = {17},
  number       = {9},
  pages        = {1103--1120},
  year         = {2007},
  url          = {https://doi.org/10.1109/TCSVT.2007.905532},
  doi          = {10.1109/TCSVT.2007.905532},
  bibsource    = {dblp computer science bibliography, https://dblp.org}
}

@inproceedings{DBLP:conf/mmm/CaoZS24,
  author       = {Qian Cao and
                  Dongdong Zhang and
                  Chengyu Sun},
  editor       = {Stevan Rudinac and
                  Alan Hanjalic and
                  Cynthia C. S. Liem and
                  Marcel Worring and
                  Bj{\"{o}}rn {\TH}{\'{o}}r J{\'{o}}nsson and
                  Bei Liu and
                  Yoko Yamakata},
  title        = {Quality Scalable Video Coding Based on Neural Representation},
  booktitle    = {MultiMedia Modeling - 30th International Conference, {MMM} 2024, Amsterdam,
                  The Netherlands, January 29 - February 2, 2024, Proceedings, Part
                  {I}},
  series       = {Lecture Notes in Computer Science},
  volume       = {14554},
  pages        = {396--409},
  publisher    = {Springer},
  year         = {2024},
  url          = {https://doi.org/10.1007/978-3-031-53305-1\_30},
  doi          = {10.1007/978-3-031-53305-1\_30},
  bibsource    = {dblp computer science bibliography, https://dblp.org}
}

@inproceedings{DBLP:conf/aaai/LuD0M24,
  author       = {Ming Lu and
                  Zhihao Duan and
                  Fengqing Zhu and
                  Zhan Ma},
  editor       = {Michael J. Wooldridge and
                  Jennifer G. Dy and
                  Sriraam Natarajan},
  title        = {Deep Hierarchical Video Compression},
  booktitle    = {Thirty-Eighth {AAAI} Conference on Artificial Intelligence, {AAAI}
                  2024, Thirty-Sixth Conference on Innovative Applications of Artificial
                  Intelligence, {IAAI} 2024, Fourteenth Symposium on Educational Advances
                  in Artificial Intelligence, {EAAI} 2014, February 20-27, 2024, Vancouver,
                  Canada},
  pages        = {8859--8867},
  publisher    = {{AAAI} Press},
  year         = {2024},
  url          = {https://doi.org/10.1609/aaai.v38i8.28733},
  doi          = {10.1609/AAAI.V38I8.28733},
  bibsource    = {dblp computer science bibliography, https://dblp.org}
}

@inproceedings{DBLP:conf/mm/WuQHLLYLY24,
  author       = {Chang Wu and
                  Guancheng Quan and
                  Gang He and
                  Xin{-}Quan Lai and
                  Yunsong Li and
                  Wenxin Yu and
                  Xianmeng Lin and
                  Cheng Yang},
  editor       = {Jianfei Cai and
                  Mohan S. Kankanhalli and
                  Balakrishnan Prabhakaran and
                  Susanne Boll and
                  Ramanathan Subramanian and
                  Liang Zheng and
                  Vivek K. Singh and
                  Pablo C{\'{e}}sar and
                  Lexing Xie and
                  Dong Xu},
  title        = {QS-NeRV: Real-Time Quality-Scalable Decoding with Neural Representation
                  for Videos},
  booktitle    = {Proceedings of the 32nd {ACM} International Conference on Multimedia,
                  {MM} 2024, Melbourne, VIC, Australia, 28 October 2024 - 1 November
                  2024},
  pages        = {2584--2592},
  publisher    = {{ACM}},
  year         = {2024},
  url          = {https://doi.org/10.1145/3664647.3680586},
  doi          = {10.1145/3664647.3680586},
  bibsource    = {dblp computer science bibliography, https://dblp.org}
}

@article{DBLP:journals/tog/KerblKLD23,
  author       = {Bernhard Kerbl and
                  Georgios Kopanas and
                  Thomas Leimk{\"{u}}hler and
                  George Drettakis},
  title        = {3D Gaussian Splatting for Real-Time Radiance Field Rendering},
  journal      = {{ACM} Trans. Graph.},
  volume       = {42},
  number       = {4},
  pages        = {139:1--139:14},
  year         = {2023},
  url          = {https://doi.org/10.1145/3592433},
  doi          = {10.1145/3592433},
  bibsource    = {dblp computer science bibliography, https://dblp.org}
}

@inproceedings{DBLP:conf/eccv/ZhangGXHWQLGZ24,
  author       = {Xinjie Zhang and
                  Xingtong Ge and
                  Tongda Xu and
                  Dailan He and
                  Yan Wang and
                  Hongwei Qin and
                  Guo Lu and
                  Jing Geng and
                  Jun Zhang},
  editor       = {Ales Leonardis and
                  Elisa Ricci and
                  Stefan Roth and
                  Olga Russakovsky and
                  Torsten Sattler and
                  G{\"{u}}l Varol},
  title        = {GaussianImage: 1000 {FPS} Image Representation and Compression by
                  2D Gaussian Splatting},
  booktitle    = {Computer Vision - {ECCV} 2024 - 18th European Conference, Milan, Italy,
                  September 29-October 4, 2024, Proceedings, Part {IX}},
  series       = {Lecture Notes in Computer Science},
  volume       = {15067},
  pages        = {327--345},
  publisher    = {Springer},
  year         = {2024},
  url          = {https://doi.org/10.1007/978-3-031-72673-6\_18},
  doi          = {10.1007/978-3-031-72673-6\_18},
  bibsource    = {dblp computer science bibliography, https://dblp.org}
}

@inproceedings{DBLP:conf/aaai/ZhuLCZJWY25,
  author       = {Lingting Zhu and
                  Guying Lin and
                  Jinnan Chen and
                  Xinjie Zhang and
                  Zhenchao Jin and
                  Zhao Wang and
                  Lequan Yu},
  editor       = {Toby Walsh and
                  Julie Shah and
                  Zico Kolter},
  title        = {Large Images Are Gaussians: High-Quality Large Image Representation
                  with Levels of 2D Gaussian Splatting},
  booktitle    = {AAAI-25, Sponsored by the Association for the Advancement of Artificial
                  Intelligence, February 25 - March 4, 2025, Philadelphia, PA, {USA}},
  pages        = {10977--10985},
  publisher    = {{AAAI} Press},
  year         = {2025},
  url          = {https://doi.org/10.1609/aaai.v39i10.33193},
  doi          = {10.1609/AAAI.V39I10.33193},
  bibsource    = {dblp computer science bibliography, https://dblp.org}
}

@inproceedings{DBLP:conf/nossdav/WangSO25,
  author       = {Longan Wang and
                  Yuang Shi and
                  Wei Tsang Ooi},
  title        = {{GSVC:} Efficient Video Representation and Compression Through 2D
                  Gaussian Splatting},
  booktitle    = {Proceedings of the 35th Workshop on Network and Operating System Support
                  for Digital Audio and Video, {NOSSDAV} 2025, Stellenbosch, South Africa,
                  31 March 2025 - 4 April 2025},
  pages        = {15--21},
  publisher    = {{ACM}},
  year         = {2025},
  url          = {https://doi.org/10.1145/3712678.3721876},
  doi          = {10.1145/3712678.3721876},
  bibsource    = {dblp computer science bibliography, https://dblp.org}
}

@inproceedings{DBLP:conf/nips/Sun0MLC024,
  author       = {Yang{-}Tian Sun and
                  Yihua Huang and
                  Lin Ma and
                  Xiaoyang Lyu and
                  Yan{-}Pei Cao and
                  Xiaojuan Qi},
  editor       = {Amir Globersons and
                  Lester Mackey and
                  Danielle Belgrave and
                  Angela Fan and
                  Ulrich Paquet and
                  Jakub M. Tomczak and
                  Cheng Zhang},
  title        = {Splatter a Video: Video Gaussian Representation for Versatile Processing},
  booktitle    = {Advances in Neural Information Processing Systems 38: Annual Conference
                  on Neural Information Processing Systems 2024, NeurIPS 2024, Vancouver,
                  BC, Canada, December 10 - 15, 2024},
  year         = {2024},
  url          = {http://papers.nips.cc/paper\_files/paper/2024/hash/5a461bdff86cc07e976bb6c518810398-Abstract-Conference.html},
  bibsource    = {dblp computer science bibliography, https://dblp.org}
}

@article{DBLP:journals/corr/abs-2411-11024,
  author       = {Weronika Smolak{-}Dyzewska and
                  Dawid Malarz and
                  Kornel Howil and
                  Jan Kaczmarczyk and
                  Marcin Mazur and
                  Przemyslaw Spurek},
  title        = {VeGaS: Video Gaussian Splatting},
  journal      = {CoRR},
  volume       = {abs/2411.11024},
  year         = {2024},
  url          = {https://doi.org/10.48550/arXiv.2411.11024},
  doi          = {10.48550/ARXIV.2411.11024},
  eprinttype    = {arXiv},
  eprint       = {2411.11024},
  bibsource    = {dblp computer science bibliography, https://dblp.org}
}

@article{DBLP:journals/corr/abs-2501-04782,
  author       = {Andrew Bond and
                  Jui{-}Hsien Wang and
                  Long Mai and
                  Erkut Erdem and
                  Aykut Erdem},
  title        = {GaussianVideo: Efficient Video Representation via Hierarchical Gaussian
                  Splatting},
  journal      = {CoRR},
  volume       = {abs/2501.04782},
  year         = {2025},
  url          = {https://doi.org/10.48550/arXiv.2501.04782},
  doi          = {10.48550/ARXIV.2501.04782},
  eprinttype    = {arXiv},
  eprint       = {2501.04782},
  bibsource    = {dblp computer science bibliography, https://dblp.org}
}

@inproceedings{DBLP:conf/cvpr/LeeCL25,
  author       = {Inseo Lee and
                  Youngyoon Choi and
                  Joonseok Lee},
  title        = {GaussianVideo: Efficient Video Representation and Compression by Gaussian
                  Splatting},
  booktitle    = {{IEEE/CVF} Conference on Computer Vision and Pattern Recognition Workshops,
                  {CVPR} Workshops 2025, Nashville, TN, USA, June 11-15, 2025},
  pages        = {4471--4480},
  publisher    = {Computer Vision Foundation / {IEEE}},
  year         = {2025},
  url          = {https://openaccess.thecvf.com/content/CVPR2025W/PBVS/html/Lee\_GaussianVideo\_Efficient\_Video\_Representation\_and\_Compression\_by\_Gaussian\_Splatting\_CVPRW\_2025\_paper.html},
  bibsource    = {dblp computer science bibliography, https://dblp.org}
}

@inproceedings{DBLP:conf/cvpr/ZhangZXLX24,
  author       = {Jiahui Zhang and
                  Fangneng Zhan and
                  Muyu Xu and
                  Shijian Lu and
                  Eric P. Xing},
  title        = {FreGS: 3D Gaussian Splatting with Progressive Frequency Regularization},
  booktitle    = {{IEEE/CVF} Conference on Computer Vision and Pattern Recognition,
                  {CVPR} 2024, Seattle, WA, USA, June 16-22, 2024},
  pages        = {21424--21433},
  publisher    = {{IEEE}},
  year         = {2024},
  url          = {https://doi.org/10.1109/CVPR52733.2024.02024},
  doi          = {10.1109/CVPR52733.2024.02024},
  bibsource    = {dblp computer science bibliography, https://dblp.org}
}

@inproceedings{DBLP:conf/3dim/ShiMGO25,
  author       = {Yuang Shi and
                  G{\'{e}}raldine Morin and
                  Simone Gasparini and
                  Wei Tsang Ooi},
  title        = {{LapisGS}: Layered Progressive {3D Gaussian} Splatting for Adaptive Streaming},
  booktitle    = {International Conference on 3D Vision, 3DV 2025, Singapore, March
                  25-28, 2025},
  pages        = {991--1000},
  publisher    = {{IEEE}},
  year         = {2025},
  url          = {https://doi.org/10.1109/3DV66043.2025.00096},
  doi          = {10.1109/3DV66043.2025.00096},
  bibsource    = {dblp computer science bibliography, https://dblp.org}
}

@inproceedings{DBLP:conf/wacv/Zoomers0MVPM25,
  author       = {Brent Zoomers and
                  Maarten Wijnants and
                  Ivan Molenaers and
                  Joni Vanherck and
                  Jeroen Put and
                  Nick Michiels},
  title        = {PRoGS: Progressive Rendering of Gaussian Splats},
  booktitle    = {{IEEE/CVF} Winter Conference on Applications of Computer Vision, {WACV}
                  2025, Tucson, AZ, USA, February 26 - March 6, 2025},
  pages        = {3118--3127},
  publisher    = {{IEEE}},
  year         = {2025},
  url          = {https://doi.org/10.1109/WACV61041.2025.00308},
  doi          = {10.1109/WACV61041.2025.00308},
  bibsource    = {dblp computer science bibliography, https://dblp.org}
}

@article{DBLP:journals/corr/abs-2503-08511,
  author       = {Yihang Chen and
                  Mengyao Li and
                  Qianyi Wu and
                  Weiyao Lin and
                  Mehrtash Harandi and
                  Jianfei Cai},
  title        = {{PCGS:} Progressive Compression of 3D Gaussian Splatting},
  journal      = {CoRR},
  volume       = {abs/2503.08511},
  year         = {2025},
  url          = {https://doi.org/10.48550/arXiv.2503.08511},
  doi          = {10.48550/ARXIV.2503.08511},
  eprinttype    = {arXiv},
  eprint       = {2503.08511},
  bibsource    = {dblp computer science bibliography, https://dblp.org}
}

@article{DBLP:journals/cacm/MildenhallSTBRN22,
  author       = {Ben Mildenhall and
                  Pratul P. Srinivasan and
                  Matthew Tancik and
                  Jonathan T. Barron and
                  Ravi Ramamoorthi and
                  Ren Ng},
  title        = {NeRF: representing scenes as neural radiance fields for view synthesis},
  journal      = {Commun. {ACM}},
  volume       = {65},
  number       = {1},
  pages        = {99--106},
  year         = {2022},
  url          = {https://doi.org/10.1145/3503250},
  doi          = {10.1145/3503250},
  bibsource    = {dblp computer science bibliography, https://dblp.org}
}

@incollection{DBLP:reference/opt/Haddad09,
  author       = {Caroline N. Haddad},
  editor       = {Christodoulos A. Floudas and
                  Panos M. Pardalos},
  title        = {Cholesky Factorization},
  booktitle    = {Encyclopedia of Optimization, Second Edition},
  pages        = {374--377},
  publisher    = {Springer},
  year         = {2009},
  url          = {https://doi.org/10.1007/978-0-387-74759-0\_67},
  doi          = {10.1007/978-0-387-74759-0\_67},
  bibsource    = {dblp computer science bibliography, https://dblp.org}
}

@inproceedings{DBLP:conf/mmsys/Shi25,
  author       = {Yuang Shi},
  editor       = {Herman Arnold Engelbrecht and
                  Silvia Rossi and
                  Simon N. B. Gunkel},
  title        = {3D Gaussian-based Immersive Media Streaming in Networked Extended
                  Reality},
  booktitle    = {Proceedings of the 16th {ACM} Multimedia Systems Conference, MMSys
                  2025, Stellenbosch, South Africa, 31 March 2025 - 4 April 2025},
  pages        = {356--360},
  publisher    = {{ACM}},
  year         = {2025},
  url          = {https://doi.org/10.1145/3712676.3719673},
  doi          = {10.1145/3712676.3719673},
  bibsource    = {dblp computer science bibliography, https://dblp.org}
}

@inproceedings{DBLP:conf/mmve/ShiZGMO24,
  author       = {Yuang Shi and
                  Ruoyu Zhao and
                  Simone Gasparini and
                  G{\'{e}}raldine Morin and
                  Wei Tsang Ooi},
  title        = {Volumetric Video Compression Through Neural-based Representation},
  booktitle    = {Proceedings of the 16th International Workshop on Immersive Mixed
                  and Virtual Environment Systems, {MMVE} 2024, Bari, Italy, April 15-18,
                  2024},
  pages        = {85--91},
  publisher    = {{ACM}},
  year         = {2024},
  url          = {https://doi.org/10.1145/3652212.3652220},
  doi          = {10.1145/3652212.3652220},
  bibsource    = {dblp computer science bibliography, https://dblp.org}
}

@article{DBLP:journals/corr/abs-2004-09576,
  author       = {Yash Bhalgat and
                  Jinwon Lee and
                  Markus Nagel and
                  Tijmen Blankevoort and
                  Nojun Kwak},
  title        = {{LSQ+:} Improving low-bit quantization through learnable offsets and
                  better initialization},
  journal      = {CoRR},
  volume       = {abs/2004.09576},
  year         = {2020},
  url          = {https://arxiv.org/abs/2004.09576},
  eprinttype    = {arXiv},
  eprint       = {2004.09576},
  bibsource    = {dblp computer science bibliography, https://dblp.org}
}

@article{DBLP:journals/taslp/ZeghidourLOST22,
  author       = {Neil Zeghidour and
                  Alejandro Luebs and
                  Ahmed Omran and
                  Jan Skoglund and
                  Marco Tagliasacchi},
  title        = {SoundStream: An End-to-End Neural Audio Codec},
  journal      = {{IEEE/ACM} Transactions on Audio, Speech, and Language Processing.},
  volume       = {30},
  pages        = {495--507},
  year         = {2022},
  url          = {https://doi.org/10.1109/TASLP.2021.3129994},
  doi          = {10.1109/TASLP.2021.3129994},
  bibsource    = {dblp computer science bibliography, https://dblp.org}
}

@article{gray1984vector,
  title={Vector quantization},
  author={Gray, Robert},
  journal={IEEE Assp Magazine},
  volume={1},
  number={2},
  pages={4--29},
  year={1984},
  publisher={IEEE}
}

@inproceedings{DBLP:conf/cvpr/AgustssonT17,
  author       = {Eirikur Agustsson and
                  Radu Timofte},
  title        = {{NTIRE} 2017 Challenge on Single Image Super-Resolution: Dataset and
                  Study},
  booktitle    = {2017 {IEEE} Conference on Computer Vision and Pattern Recognition
                  Workshops, {CVPR} Workshops 2017, Honolulu, HI, USA, July 21-26, 2017},
  pages        = {1122--1131},
  publisher    = {{IEEE} Computer Society},
  year         = {2017},
  url          = {https://doi.org/10.1109/CVPRW.2017.150},
  doi          = {10.1109/CVPRW.2017.150},
  bibsource    = {dblp computer science bibliography, https://dblp.org}
}

@inproceedings{DBLP:conf/mmsys/MercatVV20,
  author       = {Alexandre Mercat and
                  Marko Viitanen and
                  Jarno Vanne},
  editor       = {Laura Toni and
                  Ali C. Begen and
                  {\"{O}}zg{\"{u}} Alay and
                  Christian Timmerer},
  title        = {{UVG} dataset: 50/120fps 4K sequences for video codec analysis and
                  development},
  booktitle    = {Proceedings of the 11th {ACM} Multimedia Systems Conference, MMSys
                  2020, Istanbul, Turkey, June 8-11, 2020},
  pages        = {297--302},
  publisher    = {{ACM}},
  year         = {2020},
  url          = {https://doi.org/10.1145/3339825.3394937},
  doi          = {10.1145/3339825.3394937},
  bibsource    = {dblp computer science bibliography, https://dblp.org}
}

@article{DBLP:journals/corr/abs-2409-06765,
  author       = {Vickie Ye and
                  Ruilong Li and
                  Justin Kerr and
                  Matias Turkulainen and
                  Brent Yi and
                  Zhuoyang Pan and
                  Otto Seiskari and
                  Jianbo Ye and
                  Jeffrey Hu and
                  Matthew Tancik and
                  Angjoo Kanazawa},
  title        = {gsplat: An Open-Source Library for Gaussian Splatting},
  journal      = {CoRR},
  volume       = {abs/2409.06765},
  year         = {2024},
  url          = {https://doi.org/10.48550/arXiv.2409.06765},
  doi          = {10.48550/ARXIV.2409.06765},
  eprinttype    = {arXiv},
  eprint       = {2409.06765},
  bibsource    = {dblp computer science bibliography, https://dblp.org}
}

@article{DBLP:journals/pami/XieZLLY24,
  author       = {Xingyu Xie and
                  Pan Zhou and
                  Huan Li and
                  Zhouchen Lin and
                  Shuicheng Yan},
  title        = {Adan: Adaptive Nesterov Momentum Algorithm for Faster Optimizing Deep
                  Models},
  journal      = {{IEEE} Transactions on Pattern Analysis and Machine Intelligence.},
  volume       = {46},
  number       = {12},
  pages        = {9508--9520},
  year         = {2024},
  url          = {https://doi.org/10.1109/TPAMI.2024.3423382},
  doi          = {10.1109/TPAMI.2024.3423382},
  bibsource    = {dblp computer science bibliography, https://dblp.org}
}

@misc{KodakDataset,
  title        = {Kodak Lossless True Color Image Suite},
  author       = {{Eastman Kodak Company}},
  howpublished = {\url{http://r0k.us/graphics/kodak/}},
  note         = {24 images, true color, resolution $768\times512$},
  year         = {1999}
}

@inproceedings{DBLP:conf/cvpr/ZhangIESW18,
  author       = {Richard Zhang and Phillip Isola and Alexei A. Efros and Eli Shechtman and Oliver Wang},
  title        = {The Unreasonable Effectiveness of Deep Features as a Perceptual Metric},
  booktitle    = {Proceedings of the IEEE Conference on Computer Vision and Pattern Recognition (CVPR)},
  pages        = {586--595},
  year         = {2018}
}

@article{li2018vmaf,
  title={VMAF: The journey continues},
  author={Li, Zhi and Bampis, Christos and Novak, Julie and Aaron, Anne and Swanson, Kyle and Moorthy, Anush and Cock, JD},
  journal={Netflix Technology Blog},
  volume={25},
  number={1},
  year={2018}
}

@inproceedings{DBLP:conf/nips/FanWWZXW24,
  author       = {Zhiwen Fan and
                  Kevin Wang and
                  Kairun Wen and
                  Zehao Zhu and
                  Dejia Xu and
                  Zhangyang Wang},
  editor       = {Amir Globersons and
                  Lester Mackey and
                  Danielle Belgrave and
                  Angela Fan and
                  Ulrich Paquet and
                  Jakub M. Tomczak and
                  Cheng Zhang},
  title        = {LightGaussian: Unbounded 3D Gaussian Compression with 15x Reduction
                  and 200+ {FPS}},
  booktitle    = {Advances in Neural Information Processing Systems 38: Annual Conference
                  on Neural Information Processing Systems 2024, NeurIPS 2024, Vancouver,
                  BC, Canada, December 10 - 15, 2024},
  year         = {2024},
  url          = {http://papers.nips.cc/paper\_files/paper/2024/hash/fd881d3b625437354d4421818f81058f-Abstract-Conference.html},
  bibsource    = {dblp computer science bibliography, https://dblp.org}
}

@book{DBLP:books/daglib/0007442,
  author       = {David S. Taubman and
                  Michael W. Marcellin},
  title        = {{JPEG2000} - image compression fundamentals, standards and practice},
  series       = {The Kluwer international series in engineering and computer science},
  volume       = {642},
  publisher    = {Kluwer},
  year         = {2002},
  url          = {https://doi.org/10.1007/978-1-4615-0799-4},
  doi          = {10.1007/978-1-4615-0799-4},
  isbn         = {978-0-7923-7519-7},
  bibsource    = {dblp computer science bibliography, https://dblp.org}
}

@inproceedings{DBLP:conf/wacv/PrestaTFGC25,
  author       = {Alberto Presta and
                  Enzo Tartaglione and
                  Attilio Fiandrotti and
                  Marco Grangetto and
                  Pamela C. Cosman},
  title        = {Efficient Progressive Image Compression with Variance-Aware Masking},
  booktitle    = {{IEEE/CVF} Winter Conference on Applications of Computer Vision, {WACV}
                  2025, Tucson, AZ, USA, February 26 - March 6, 2025},
  pages        = {7692--7700},
  publisher    = {{IEEE}},
  year         = {2025},
  url          = {https://doi.org/10.1109/WACV61041.2025.00747},
  doi          = {10.1109/WACV61041.2025.00747},
  bibsource    = {dblp computer science bibliography, https://dblp.org}
}

@article{DBLP:journals/tomccap/ZhangZT24,
  author       = {Gai Zhang and
                  Xinfeng Zhang and
                  Lv Tang},
  title        = {Unified and Scalable Deep Image Compression Framework for Human and
                  Machine},
  journal      = {{ACM} Trans. Multim. Comput. Commun. Appl.},
  volume       = {20},
  number       = {10},
  pages        = {314:1--314:22},
  year         = {2024},
  url          = {https://doi.org/10.1145/3678472},
  doi          = {10.1145/3678472},
  bibsource    = {dblp computer science bibliography, https://dblp.org}
}

@inproceedings{zhao2024scaling,
  title={On scaling up 3d gaussian splatting training},
  author={Zhao, Hexu and Weng, Haoyang and Lu, Daohan and Li, Ang and Li, Jinyang and Panda, Aurojit and Xie, Saining},
  booktitle={European Conference on Computer Vision},
  pages={14--36},
  year={2024},
  organization={Springer}
}

@standard{ISO13818-2,
  title        = {Information technology --- Generic coding of moving pictures and associated audio information --- Part 2: Video},
  organization = {International Organization for Standardization (ISO) and International Electrotechnical Commission (IEC)},
  number       = {ISO/IEC 13818-2},
  year         = {2000},
  note         = {Also published as ITU-T Recommendation H.262}
}

@standard{ISO14496-2,
  title        = {Information technology --- Coding of audio-visual objects --- Part 2: Visual},
  organization = {International Organization for Standardization (ISO) and International Electrotechnical Commission (IEC)},
  number       = {ISO/IEC 14496-2},
  year         = {2004}
}

@article{DBLP:journals/corr/abs-2601-05394,
  author       = {Yuang Shi and
                  G{\'{e}}raldine Morin and
                  Simone Gasparini and
                  Wei Tsang Ooi},
  title        = {Sketch{\&}Patch++: Efficient Structure-Aware 3D Gaussian Representation},
  journal      = {CoRR},
  volume       = {abs/2601.05394},
  year         = {2026},
  url          = {https://doi.org/10.48550/arXiv.2601.05394},
  doi          = {10.48550/ARXIV.2601.05394},
  eprinttype    = {arXiv},
  eprint       = {2601.05394},
  bibsource    = {dblp computer science bibliography, https://dblp.org}
}

@inproceedings{sun2025tsla,
  author    = {Sun, Yuan-Chun and Shi, Yuang and Lee, Cheng-Tse and Zhu, Mufeng and Ooi, Wei Tsang and Liu, Yao and Huang, Chun-Ying and Hsu, Cheng-Hsin},
  title     = {{LTS}: A {DASH} Streaming System for Dynamic Multi-Layer {3D Gaussian} Splatting Scenes},
  publisher = {{ACM}},
  booktitle={Proceedings of the 16th {ACM} Multimedia Systems Conference, MMSys 2025, Stellenbosch, South Africa, March 31-April 4, 2025},
  year      = {2025},
  url       = {https://doi.org/10.1145/3712676.3714445},
  doi       = {10.1145/3712676.3714445},
}

\end{document}